\documentclass[10pt]{article}
\usepackage[margin=1in]{geometry}
\usepackage[sc]{mathpazo}
\usepackage[ruled]{algorithm2e}
\usepackage{graphicx}  
\usepackage{amsfonts}
\usepackage{amsmath} 
\usepackage{amssymb}
\usepackage{mathtools}
\usepackage{booktabs}
\usepackage{siunitx}
\usepackage[x11names]{xcolor}
\usepackage[authoryear,round]{natbib} 
\usepackage{tikz}
\usepackage{multirow}
\usepackage{comment}
\usepackage[many]{tcolorbox}
\usepackage{setspace}
\usepackage{multicol}
\usepackage{dsfont}
\usepackage{fancyhdr}
\usepackage{etoolbox}
\usepackage[dvipsnames,x11names,svgnames,table]{xcolor}

\makeatletter

\renewcommand{\@dotsep}{10000} 
\providecommand{\l@subparagraph}[2]{}
\makeatother

\definecolor{BlueGreen}{HTML}{009F6B}
\usepackage[colorlinks=true, linkcolor=BlueGreen, citecolor=BlueGreen, urlcolor=BlueGreen]{hyperref}

\newtcolorbox{boxE}{
  enhanced, breakable,            
  colback=white,        
  colframe=Orange,   
  boxrule=0.75pt,       
  arc=0pt,              
  left=6pt, right=6pt, top=6pt, bottom=6pt
}

\allowdisplaybreaks

\title{Inexact calculus of variations on the hyperspherical tangent bundle with connections to the attention mechanism}

\author{
  Andrew Gracyk 
}
\date{}

\begin{document}

\maketitle

\begin{abstract}
\noindent We offer a theoretical mathematical background through Lagrangian optimization on the unit hyperspherical manifold and its tangential structure. Our methods can be categorized as inexact since our methods are projection-based and since we will perturb the functional optimization with epsilon-type quantities. We draw connections to the attention mechanism and the Transformer since it exists as a flow map in the tangent fiber for each token along the high-dimensional unit sphere. Our motivation for this work is primarily twofold: we study the attention mechanism under its flow map and its relations to traditional calculus of variations and Lagrangian optimization; and we study several calculus problems of variations on the unit hypersphere that appeal to a broader mathematical lens in approximating, variational contexts.
\end{abstract}


\vspace{2mm}

\textbf{AMS MSC Classifications (2020):} 68T07, 49K15, 49Q99, 49J15

\setcounter{tocdepth}{2}
\tableofcontents

\section{Preliminaries}
\label{sec:preliminaries}

We will investigate two paradigms in this work and attempt to reconcile them. The attention mechanism inherently satisfies a manifold constraint as well as a dynamic flow map under empirical conditions, therefore we study how this mechanism relate to classical calculus of variations of manifolds. Moreover, we study a broader calculus of variations lens for more generalized mathematical purposes while maintaining faithfulness to dynamics on the unit hypersphere. This mathematical approach will slightly deviate from typical literature because our methods will be manifold-based but not along the manifold definitively. As we will see in this work here, the Transformer satisfies a pre-determined flow map along the unit hyperspherical manifold under a projection mapping, but the starting token point is permitted freedom. The traditional Euler-Lagrange equation is inverted: the flow map has freedom, but the initial and terminal points are fixed.

\vspace{2mm}

Our starting point for this Transformer aspect of work is that of \cite{geshkovski2024mathematicalperspectivetransformers}. This work discusses the Transformer as it exists as a flow map on the unit hypersphere. A \textit{token} is a sequence element vector $x_i(0) \in \mathbb{R}^d$, which is a latent vector in which the input data, i.e. a collection of text in a large language model, is mapped. The Transformers exists on the unit hypersphere due to each token is divided by its Euclidean norm, and a diagonal matrix is multiplied. The work of \cite{geshkovski2024mathematicalperspectivetransformers} works under the assumption this trained diagonal matrix is the identity, so we will operate under this crucial assumption as well. We refer to \cite{geshkovski2024mathematicalperspectivetransformers} for more details.

\vspace{2mm}

The existence of particle trajectories along this manifold have a natural connection to calculus of variations in the context of functional optimization. Calculus of variations \cite{krupka2015global} \cite{doi:10.1142/2199} \cite{musilova2025fibred} is the study of functionals and their variations, i.e. it is the calculus analog for functional-based quantities. Oftentimes, this broader mathematical area involves optimizing sufficiently nice functions such as smooth or possibly convex Lagrangians among various paths and trajectories. The classical Euler-Lagrange equation optimizes a path among an initial and terminal endpoint.

\vspace{2mm}

Calculus of variations on manifolds is frequently done intrinsically \cite{krupka2001geometricaspectsvariationalproblems}, while our approach is not so. The flow map of the Transformer is done via a projection onto the tangent fiber at each token, and so the flow map in the tangent space is not automatic. Thus, our methods reflect both extrinsic and intrinsic geometric calculus of variations, primarily through the use of such a projection mapping. Moreover, such a flow map has a very particular closed form which is highly nonstandard is calculus of variations literature. However, this is not to say that the intrinsic and extrinsic mixture approach to manifold-based calculus of variations is new. Indeed, flow maps in tangent spaces are holonomic \cite{J_F_Carinena_1993} \cite{Lemos_2021} since the projections only rely on position and time, thus these calculus of variations scenarios are well-studied and follow the quintessential Euler-Lagrange equation. We outline our contributions.

\vspace{2mm}

We provide a specific Lagrangian that all Transformers satisfy up to layer count and normalization. Moreover, we develop loss minimization scenarios in which our methods are compatible. Moreover, these contexts are more generalized than just to Transformers.

\vspace{2mm}

We provide a fundamental calculus of variations result inherent to the Transformer. We discuss the projected Euler Lagrange equation on the unit hypersphere, which is a generalized result. We remark the setups of these two problems differ.

\vspace{2mm}

We provide an adaptation of the Euler-Lagrange up to an infimum-satisfaction with an added error tolerance. We show the set that satisfies this has nonzero Hausdorff measure, which is arguably trivial due to smoothness. Moreover, our primary result in this regard is showing the manifold Hausdorff measure of this set is bounded below by the measure of a sufficiently large geodesic ball. We provide a subsidiary result based on this proof specific to the Transformer.

\vspace{2mm}

We prove a discretized Transformer flow map approaches the continuous version, i.e. the integral, under the condition that the time discretization is sufficiently small. For the Transformer, the time discretization getting small is equivalent to the number of layers increasing \cite{geshkovski2024mathematicalperspectivetransformers}, thus this result is useful.

\vspace{2mm}

We provide a result showing that a pushforward of measures to minimize a Lagrangian satisfies the calculus of variations problem under Dirac point masses. 

\vspace{2mm}

Lastly, we provide a kinetic energy result based on geodesics. This result is compatible with an Euler-Lagrange setup, as opposed to a Transformer setup, so the endpoints are fixed but the path is not.

\section{The Transformer as a flow map}

We discuss the preliminary mathematical foundations of the Transformer \cite{geshkovski2024mathematicalperspectivetransformers}, its dynamical evolution, and its relations to manifold geometry \cite{LIU2019168}. The introduction of the attention mechanism of \cite{vaswani2023attentionneed} ushered in a new epoch of machine learning. We will be studying a variant of this equation under the rough equivalence
\begin{align}
\label{eqn:attention}
\text{Attention}(Q,K,V)_i = \sum_j \frac{ \exp ( q_i \cdot k_j / \sqrt{d_j} ) }{ \sum_l \exp ( q_i \cdot k_l / \sqrt{d_l} ) } v_j \propto \mathcal{P}_{x_i(t)}^{\perp} \Bigg(  \frac{1}{Z_{\beta,i}(t)}   \sum_{j=1}^n \Big( e^{ \beta  \langle Q(t) x_i(t), K(t) x_j(t) \rangle} V(t) x_j(t) \Big) \Bigg)  .
\end{align}
Here, $x_i(t)$ is the token as a function of time. $x_i(0)$ is the initial token, and the time-dependence is the continuum limit of the data as it is processed sequentially via layers \cite{geshkovski2024mathematicalperspectivetransformers}. $Z$ is defined as in \ref{eqn:flow_map_discrete}. It is not the case that the token evolves in time because the ambient data evolves in time, which is an easy misconception. It is by the construction of the neural network layers, and by constructing a continuous analog.

\vspace{2mm}

The projection is an important caveat. The attention mechanism itself has no manifold projection mapping, so the equation of \ref{eqn:attention} is perhaps a bit misleading. The projection takes place because it is intended to mimic the effect of a layer normalization \cite{geshkovski2024mathematicalperspectivetransformers} in a discrete setting. It is not reasonable to attempt to normalize in an infinite-depth setting, so the projection operator is a clear semblance to what the layer normalization attempts to induce among the data processing.

\vspace{2mm}

For the remainder of this manuscript, we will refer to a \textit{token} as
\begin{align}
\text{token} = x_i(0) \in (x_i(0))_{i \in [n]} \in \mathbb{R}^d .
\end{align}
The token refers to the initial embedding only, i.e. the initial data at layer zero. It is not the time-dependent version of the positioning along the sphere. We deviate from the terminology of \cite{geshkovski2024mathematicalperspectivetransformers} a bit. We refer to $\{x_i(t)\}_i$ as particles, trajectories, paths, or token trajectories, which are the ODE-evolved versions of the tokens with the token as the starting position. As an additional remark, note that we will investigate the sphere $\mathbb{S}^{d-1}$ embedded in $\mathbb{R}^d$. This construction is that of a hypersurface, which is notable because not all (if not most) manifolds are embedded in extrinsic space exactly one dimension higher than its intrinsic.

\vspace{2mm}

We discuss the role of $\beta$. Here, $\beta > 0$ is a fixed constant that is inherent to the particular model. In the language of statistical physics, $\beta$ corresponds to notions of inverse temperature \cite{geshkovski2024mathematicalperspectivetransformers}. Our investigation is more mathematical in nature rather than pertaining to physics, so we will solely treat $\beta$ as a positive real number rather than investigate its physical significance. Generally, in Transformers, $\beta$ is taken quite small. By equation \ref{eqn:attention}, $\beta$ has relevance to $\sqrt{d}$. This is quite notable, as some of our analysis will depends on the orders of $\beta$. For example, Laplace's method is a technique to approximate an integral of a form quite similar to that in \ref{eqn:attention}, but this approximation is near nonsense for us because it relies on $\beta$ quite large.

\vspace{2mm}

We will be studying the transformer as flow map. Our trajectories effectively evolve under the ODE
\begin{align}
\text{evolving token at} \ t = \text{token at} \ 0 + \int_0^t \mathcal{X}[\mu](\text{evolving token}) dt .
\end{align}
We have defined $\mathcal{X}$ in equation \ref{eqn:flow_map_cont}. As a flow map, the Tranformer architecture satisfies the dynamical evolution
\begin{align}
\label{eqn:flow_map_discrete}
\dot{x}_i (t) & = \mathcal{P}_{x_i(t)}^{\perp} \Bigg(  \frac{1}{Z_{\beta,i}(t)}   \sum_{j=1}^n \Big( e^{ \beta  \langle Q(t) x_i(t), K(t) x_j(t) \rangle} V(t) x_j(t) \Big) \Bigg) 
\\
& = \mathcal{P}_{x_i(t)}^{\perp} \Bigg(  \frac{1}{ \underbrace{ \sum_{j=1}^n  e^{ \beta  \langle Q(t) x_i(t), K(t) x_j(t) \rangle} }_{= Z_{\beta,i}(t)}}   \sum_{j=1}^n \Big( e^{ \beta  \langle Q(t) x_i(t), K(t) x_j(t) \rangle} V(t) x_j(t)  \Big) \Bigg)  .
\end{align}
Here, $Z$ is the dividing normalization factor, and $t \geq 0$ is a time-parameter used to denote the continuous-analog of output data under a time discretization, and $x_i \in \mathbb{S}^{d-1}$. We refer to \cite{geshkovski2024mathematicalperspectivetransformers}, page 5, for a thorough explanation as to why $x_i \in \mathbb{S}^{d-1}$ is valid. We have defined the projection operator
\begin{align}
\mathcal{P}_p^{\perp}(y)  : \mathbb{R}^d \rightarrow T_p \mathbb{S}^{d-1} = y - \Big\langle y, p \Big\rangle p .
\end{align}
We have used notation $T_p \mathbb{S}^{d-1}$ to denote the tangent space at point $p$ of the unit hypersphere $\mathbb{S}^{d-1}$, i.e.
\begin{align}
T_p \mathbb{S}^{d-1} = \text{span} \Big\{ \frac{\partial}{\partial u^1} \Big|_p, \hdots, \frac{\partial}{\partial u^{d-1}} \Big|_p \Big\} .
\end{align}
As a remark, we will mathematically show that $\mathcal{P}_p^{\perp}$ is indeed in the tangent space of the hypersphere, since it is critical to our work. Observe
\begin{align}
\langle y - \langle y,p \rangle p, p \rangle = \langle y,p \rangle - ||p||^2 \langle y, p \rangle = 0 ,
\end{align}
since $||p||=1$ due to the assumption. In particular, we have each hyperplane is orthogonal to the vector along the hypersphere.

\vspace{2mm}

This flow map has a complementary measure-theoretic formulation
\begin{align}
\label{eqn:flow_map_cont}
\dot{x}_i(t) = \mathcal{X}[\mu(t)](x_i(t)) & = \mathcal{P}_{x_i(t)}^{\perp} \Bigg( \frac{1}{Z_{\beta,\mu}(x_i(t))} \int e^{\beta \langle x_i(t),y \rangle} y d\mu(t,y) \Bigg) \\
& = \mathcal{P}_{x_i(t)}^{\perp} \Bigg( \frac{1}{\underbrace{\int e^{\beta \langle x_i(t), y \rangle} d\mu(t,y)}_{= Z_{\beta, \mu}(x(t))}  } \int e^{\beta \langle x_i(t),y \rangle} y d\mu(t,y) \Bigg) .
\end{align}
Here,
\begin{align}
\mu(t,\cdot) = \frac{1}{n}   \sum_i \delta_{x_i(t)}  ( \cdot ) 
\end{align}
is the empirical measure across the trajectories. Note that this measure satisfies a continuity equation \cite{geshkovski2024mathematicalperspectivetransformers}. In our analysis, we will generally work with this measure-theoretic form, since it is more generalized than the discrete formulation, but they are equivalent up to the continuum limit.

\section{Geometry background}

Since we have outlined that the dynamics of $\dot{x}_i(t)$ lie in the collective tangent spaces of the hypersphere, we will turn our attention to some geometry background on the tangent space and the tangent bundle.

\vspace{2mm}

First, we discuss the tangent bundle \cite{tangent_bundle_nlab} \cite{rowland2025tangentbundle}. The tangent bundle of the sphere $\mathbb{S}^{d-1}$ is the space
\begin{align}
T \mathbb{S}^{d-1} & = \bigsqcup_{p \in \mathbb{S}^{d-1} } T_p \mathbb{S}^{d-1} = \Big\{ (p,y) | p \in \mathbb{S}^{d-1} , y \in T_p \mathbb{S}^{d-1}  \Big\} .
\end{align}
It is the totality of all the tangent spaces across all points $p$. The tangent bundle is the corresponding pair; it is not just the tangent space union. Defining the projection map and the fiber with
\begin{align}
\pi : T \mathbb{S}^{d-1} \rightarrow \mathbb{S}^{d-1}, \pi^{-1}(p) = T_p \mathbb{S}^{d-1} ,
\end{align}
we get the bundle
\begin{align}
( T \mathbb{S}^{d-1}, \pi, \mathbb{S}^{d-1}) .
\end{align}
Let $d$ denote the intrinsic dimension of a smooth Riemannian manifold $(\mathcal{M},g)$. Let $\Psi$ be the diffeomorphism
\begin{align}
\Psi : \Sigma \subseteq \mathcal{M} \xrightarrow{\sim}\mathbb{R}^{d-1}, \mathcal{M} \cong \mathbb{R}^{d-1} ,
\end{align}
so that the manifold is diffeomorphic to $d-1$-dimensional Euclidean space. Note that each point in the tangent bundle can be written using the tangent basis, i.e.
\begin{align}
\Big( \Psi \in T\mathcal{M} = \bigsqcup_p T_p \mathcal{M} \Big) \in \Bigg\{ \sum_{i=1}^d X^i \frac{\partial}{\partial u^i} \Bigg|_p : X^i \in \mathbb{R} \Bigg\} .
\end{align}
Let us return to our spherical context. First, we note that the manifold
\begin{align}
\pi_{\mathbb{S}} : \mathbb{R}^d \setminus \{0\} \to \mathbb{S}^{d-1}, \quad \psi \mapsto \frac{\psi}{\|\psi\|_2}
\end{align}
is a $(d-1)$-dimensional manifold embedded in $\mathbb{R}^d$, with the standard inclusion map $\iota : \mathbb{S}^{d-1} \hookrightarrow \mathbb{R}^d$.

\vspace{2mm}

Our measures in our arguments will be taken with respect to the manifold in a sufficiently small ball-type area around the starting token $x_i(0)$, specifically at $t=0$. Our measure theoretic-integrals will be taken specifically at $t=0$, thus we will be integrating along this ball along a collection of trajectories but just at the starting time. We will work with Hausdorff measure (up to constants)
\begin{align}
\mathcal{H}^k(\Sigma) = \liminf_{\delta \rightarrow 0} \Bigg\{ \sum_i ( \text{diam}(\Sigma_i))^k : \Sigma \subset \bigcup_i \Sigma_i, \text{diam}(\Sigma_i) < \delta \Bigg\} .
\end{align}
Note that a volume element is typically done via a metric, which is dependent on local coordinates. Hausdorff measure has no need for local coordinates, and there is a known equivalence on manifolds
\begin{align}
\mathcal{H} = \mu ,
\end{align}
where $\mu$ is Lebesgue measure, so without loss of generality this choice of measure is not of extreme importance.

\vspace{2mm}

Specifically, our measure-theoretic integrals will take the form
\begin{align}
\int_{y(0) \in \Sigma \subseteq \mathbb{S}^{d-1}} f(y,t) d\mathcal{H}^k(y(0))   .
\end{align}
We will make the reasonable assumptions necessary to assume this is well-defined. To illustrate clearly, we provide a discrete analog: we have
\begin{align}
\int_{y(0) \in \Sigma \subseteq \mathbb{S}^{d-1}} f(y,t) d\mathcal{H}^k(y(0))   \approx \lim_{\max \text{diam}(E_j) \rightarrow 0} \sum_{j=1}^N f(y_j,t) \mathcal{H}^k(E_j) .
\end{align}
Here $\{E_j\}_j$ is a partition of $\Sigma$. We sum up all of the paths that start in $\Sigma$, thus the function after integration is still time-dependent, since we are only summing up the pathways based on their initial starting points. In particular, we will analyze functions that evolve according to well-posed ODEs. We define the flow $\Psi_t : y_0 \rightarrow y(t;y_0)$. Our analysis will consist of taking integrals of the form
\begin{align}
\int_0^T \int_{y_0 \in \Sigma} f(t,\Psi_t(y_0)) d\mu(y_0) dt = \int_0^T \int_{y_0 \in \Sigma} f(t,y(t,y_0)) d\mu(y_0) dt.
\end{align}
We will assume basic regularity facts such as measurability (in fact, we will generally work with smoothness to sufficient order).

\vspace{2mm}

In some of our work, we use the Riemannian exponential map. The reason we do this is so have a well-defined mapping from a perturbation around the critical path to the manifold. The exponential map is map from the tangent space of the manifold to the manifold itself via
\begin{align}
h \in T_{x} \mathbb{S}^{d-1} \rightarrow \exp_x(h) \in \mathbb{S}^{d-1} .
\end{align}
We effectively begin at a base point $x$ and move along $h$ as a geodesic to arrive at new vector $\exp_x(h)$. The exponential map is primarily intrinsically motivated, but since this output a point on the hypersphere for us, it also has extrinsic connections.

\section{Lagrangian background}
\label{sec:Lagrangian_background}

The primary goal of ours will be an attempt to study the calculus of variations problem
\begin{align}
\begin{cases}
\min_{x(0) \ \text{is a token}} \int_0^T L(t,x(t),\dot{x}(t)) dt
\\
\text{subject to manifold constraints} .
\end{cases}
\end{align}
However, we will study this problem up to various levels of extremity, since this calculus of variations problem is highly generalized. Note that the flow map of the Transformer is deterministic. The ODE that governs the evolution of each token is fixed, up to the training parameters. Hence, our optimization will primary be done such that the starting point of each ODE is learned, rather than learning functionals along arbitrary trajectories with initial and terminal constraints.

\textbf{\begin{figure}[t]
  \vspace{0mm}
  \centering
  \includegraphics[scale=0.7]{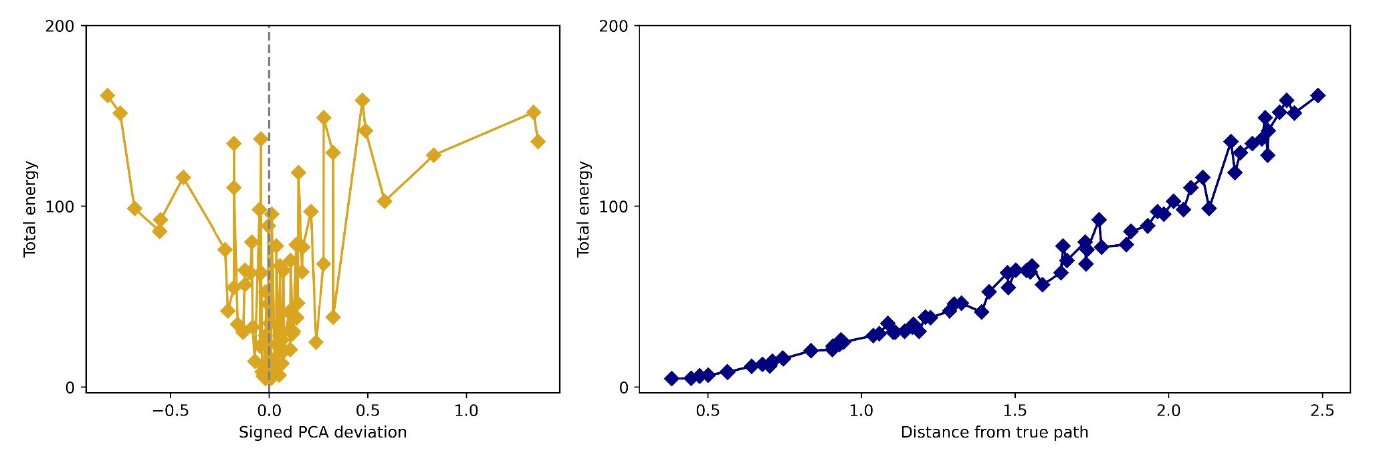}
  \caption{We plot (left) the signed PCA deviation and (right) the energy of the functional against the $L^2$ discrete distance from the true trajectory as in Theorem 1. We use a low dimension here $(d=3)$. The number of points corresponds to the number of trials of $h$. Here, $h$ is a test trajectory that was deliberately constructed and fixed instead of using the flow map of \ref{eqn:flow_map_cont}, thus this figure is primarily illustrative regarding Theorem 1. $h$ is constructed by injecting noise along the true path then projecting onto the sphere. The signed PCA deviation corresponds to the direction of deviation of the variational path from the true path. As we can see, $[\dot{h} = \text{true path derivative}]$ is optimal.}
\label{fig:energy_thm1_lowdim}
\end{figure}}

\vspace{2mm}

We will sometimes work with specific Lagrangians. One particular very classical Lagrangian is the geodesic equation
\begin{align}
\int_0^T g( \dot{x}, \dot{x} ) dt = \int_0^T g_{ij} \dot{x}^i \dot{x}^j dt ,
\end{align}
whose solution is the geodesic equation. We will also work with the kinetic-potential energy
\begin{align}
L = \| \dot{x} \|_2^2 +  \log ( \int e^{\beta \langle x(t), y \rangle} d\mu(t,y) ) .
\end{align}
This choice of Lagrangian is quite natural for the Transformer, since the same integral appears in both the first derivative of the particles as well as the potential energy
\begin{align}
V = \log ( \int e^{\beta \langle x(t), y \rangle} d\mu(t,y) ) .
\end{align}
In particular, we have the Transformer flow map is a gradient flow of the projected potential energy $V$ with respect to the Euclidean metric, i.e.
\begin{align}
\dot{x}(t) = \frac{1}{\beta}  \mathcal{P}_{x(t)}^{\perp} ( \nabla V(x(t)) ) = \mathcal{X}[\mu(t)](x(t)) .
\end{align}
This result is also briefly discussed in \cite{geshkovski2024mathematicalperspectivetransformers}, but outside the context of Lagrangians. Notice
\begin{align}
& \frac{1}{\beta} \mathcal{P}_{x(t)}^{\perp} ( \nabla_x V(x(t)) ) = \frac{1}{\beta}  \mathcal{P}_{x(t)}^{\perp} (  \nabla_x \log ( \int e^{\beta \langle x(t), y \rangle} d\mu(t,y) ) ) 
\\
= \ & \frac{1}{\beta}  \mathcal{P}_{x(t)}^{\perp} (  \frac{1}{Z_{\beta}(x(t))} \int \beta y e^{\beta \langle x(t), y \rangle} d\mu(t,y) ) = \mathcal{X}[\mu(t)](x(t))
\end{align}

\vspace{2mm}

What we have is the derivative field of the potential produces exactly the velocity vector field of the particles. This is quite a normal condition in many gradient flow-based physical systems 
\begin{align}
v = - \nabla V ,
\end{align}
thus this choice of Lagrangian is sensible for our work. In particular, we have chosen our Lagrangian such that the Transformer trajectories flow in the direction of steepest decent of potential energy. Note that this formulation utilizes the baseline Euclidean metric and not the Wasserstein metric, which is common in literature. For discussion of Wasserstein gradient flows of Transformers, we refer to \cite{geshkovski2024mathematicalperspectivetransformers}. This formulation represents how the tokens move across time to order to minimize such an energy. In our analysis, we will use the potential
\begin{align}
\Phi = \Phi(x(t)) = \log ( \int e^{\beta \langle x(t), y \rangle} d\mu(t,y) ) .
\end{align}

\vspace{2mm}

The gradient flow property of this Lagrangian functional has significance. We know that a token's trajectory never increases its energy (at least its potential energy). The tokens are initialized somehow according to the training optimization, then they proceed to move downhill according to the potential energy problem. More specifically, the optimal solution satisfies
\begin{align}
\delta ( \int_0^T   \| \dot{x} \|_2^2 +  \log ( \int e^{\beta \langle x(t), y \rangle} d\mu(t,y) )  dt ) = 0 .
\end{align}
Considering a test function added $\epsilon h(t)$, and using integration by parts with suitable boundary conditions on $h$, we notice
\begin{align}
0 = \frac{d}{d \epsilon} L(x + \epsilon h) |_{\epsilon = 0} & = \int_0^T \frac{d}{d \epsilon }  ( \| \dot{x} + \epsilon \dot{h} \|_2^2 +  \log ( \int e^{\beta \langle x(t) + \epsilon h(t), y \rangle} d\mu(t,y) ) ) \Bigg|_{\epsilon = 0}  dt 
\\
& = \int_0^T ( 2 \langle \dot{x}(t), \dot{h}(t) \rangle + \langle \nabla V(x(t)), h(t) \rangle ) dt
\\
& = \int_0^T \langle -2 \ddot{x}(t) + \nabla V(x(t)), h(t) \rangle dt \implies \mathcal{P}_{x(t)}^{\perp} ( -2 \ddot{x}(t) + \nabla V(x(t)) ) = 0 .
\end{align}
Note that this is exactly a projected Euler-Lagrange equation. This is equivalent to exactly Newton's second law
\begin{align}
a = -\frac{1}{m} \nabla U(x)
\end{align}
for potential $U$. Thus, this particular choice of Lagrangian has broad applications in physics.

\vspace{2mm}

We return to discuss our functional choice in section \ref{sec:preliminaries}. Recall the functional we introduced there. Using the exact potential we introduced, we present:

\vspace{2mm}

\textbf{Theorem 1.} \textit{The trajectory of every token of the Transformer, under our assumptions, satisfies an Euler-Lagrange equation on the unit hypersphere with respect to the functional
\begin{align}
\label{eqn:transformer_functional}
\int_0^T (  \frac{1}{2} \|\dot{h}(t)\|_2^2 - \langle \dot{h}(t), \mathcal{P}_{x(t)}^{\perp} ( \nabla_x \log ( \int e^{\beta \langle x(t), y \rangle} d\mu(t,y) )  ) \rangle + \frac{1}{2T} \| h(0) - x(0) \|_2^2 ) dt ,
\end{align}
in the continuum limit, where the optimization runs over $h$, and $x(t)$ is a fixed vector that represents the token along its trajectory in its embedding at $t$.
}

\vspace{2mm}

Note that this Theorem is actually quite generalized, and can be applied to a wide range of calculus of variations problems where the flow map of the first derivative follows a closed form. This Theorem is instantiated to the Transformer through the exact potential used.

\vspace{2mm}

We will study the Euler-Lagrange equation more closely in section \ref{sec:euler-lagrange}. Observe this functional achieves a critical point at the Euler Lagrange equation
\begin{align}
0 = \ & \frac{\partial L}{\partial h} - \frac{d}{dt} ( \frac{\partial L}{\partial \dot{h}} ) =  \frac{d}{dt} ( \dot{h} - \mathcal{P}_{x(t)}^{\perp} ( \nabla_x \log ( \int e^{\beta \langle x(t), y \rangle} d\mu(t,y) )  )  ) 
\end{align}
Note that this implies the difference is equal to some constant of integration. In particular, we get
\begin{align}
\dot{h} - \mathcal{P}_{x(t)}^{\perp} ( \nabla_x \log ( \int e^{\beta \langle x(t), y \rangle} d\mu(t,y) )  ) = \text{constant} .
\end{align}
We will show this constant gets annihilated with a first variation argument, since the boundary is arbitrary. In particular, the first variation condition necessitates every admissible $h$ satisfies what is desired. Consider
\begin{align}
\delta \mathcal{A}[h,\delta h] = \int_0^T \langle \dot{h}, \delta \dot{h} \rangle - \langle \delta \dot{h}, \mathcal{P}^{\perp} ( \nabla \Phi) \rangle dt = \int_0^T \langle \dot{h} - \mathcal{P}^{\perp} ( \nabla \Phi), \delta \dot{h} \rangle dt .
\end{align} 
After integration by parts,
\begin{align}
\delta \mathcal{A} = 0 = \langle \dot{h} - \mathcal{P}^{\perp} ( \nabla \Phi), \delta h \rangle \Big|_{t =0}^{t = T} - \int_0^T \langle \frac{d}{dt} ( \dot{h} - \mathcal{P}^{\perp} ( \nabla \Phi) ), \delta h \rangle dt .
\end{align}
We must have
\begin{align}
& \langle (\dot{h} - \mathcal{P}^{\perp} ( \nabla \Phi))|_{t=T}, \delta h(T) \rangle - \langle (\dot{h} - \mathcal{P}^{\perp} ( \nabla \Phi))|_{t=0}, \delta h(0) \rangle
\\
& = \langle \text{constant}, \delta h(T) \rangle - \langle \text{constant}, \delta h(0) \rangle = 0,
\end{align}
(recall this is the case since first variation is zero at a critical point, and recall the integral is zero by the Euler-Lagrange equation). Since the boundary variation endpoints are free, this implies the constant is zero. This is a crucial remark: for this argument, we have used Neumann boundary conditions. This is important because we will sometimes use Dirichlet boundary conditions. Thus, every Transformer implicitly satisfies a critical point of this functional via an Euler-Lagrange equation.

\vspace{2mm}

It is worth considering changing $x$ to the variational path $h$ in the formulation of \ref{eqn:transformer_functional}, but this now becomes an arbitrary flow map and does not correspond to an actual token, thus it loses meaning. For this purpose, we maintain the use of $x$ in this functional.

\vspace{2mm}

Here, we have let $h$ be unrestricted in Euclidean space. Alternatively, we can slightly adjust the problem of \ref{eqn:transformer_functional} under a constraint, which is arguable also more aligned with a typical Transformer. 

\vspace{2mm}

Note that \ref{eqn:attention} is not exactly satisfied in the discrete setting, which is the typical setting in which the real Transformer exists. Our methods are primarily a theoretical framework in which to analyze the Transformer, but the Transformer more accurately follows the discrete setting over the continuous one. In particular, a more practical Lagrangian functional is of the surrogate class
\begin{align}
\widetilde{\mathcal{A}} = \sum_{t} \Bigg[ \frac{1}{2} \|h_t\|_2^2 - \langle h_t, \mathcal{P}_{x_t}^{\perp}(\nabla_x \Phi(x_t)) \rangle \Bigg] \Delta t .
\end{align}
This is only an approximation to the true functional, thus satisfaction of this equation in a first variation-type argument is only approximate. The theoretical functional minimization framework provides an analytic risk argument to solving the flow map, not an actual one. The discrete paradigm provides us a risk framework to understand the underlying variational dynamics, but not the ones faithful to how it actually performs. 

\vspace{2mm}

Therefore, in much of our analysis, we will operate so that the true (the discrete) functional is satisfied approximately, i.e. the Euler-Lagrange equation of $\widetilde{A}$ is satisfied exactly (although this is approximated as well in actuality, but this error is absorbed into the other error). Thus, we will analyze
\begin{align}
h \ \text{solves} \ \int L(h,\dot{h}) dt \leq \min_x \int L(x,\dot{x}) dt + \epsilon ,
\end{align}
and so the continuum-based functional is estimated. Note that the solution to this is an approximation to $\mathcal{P}_x^{\perp} ( \nabla \Phi)$ assuming that $h$ is unconstrained.

\textbf{\begin{figure}[t]
  \vspace{0mm}
  \centering
  \includegraphics[scale=0.7]{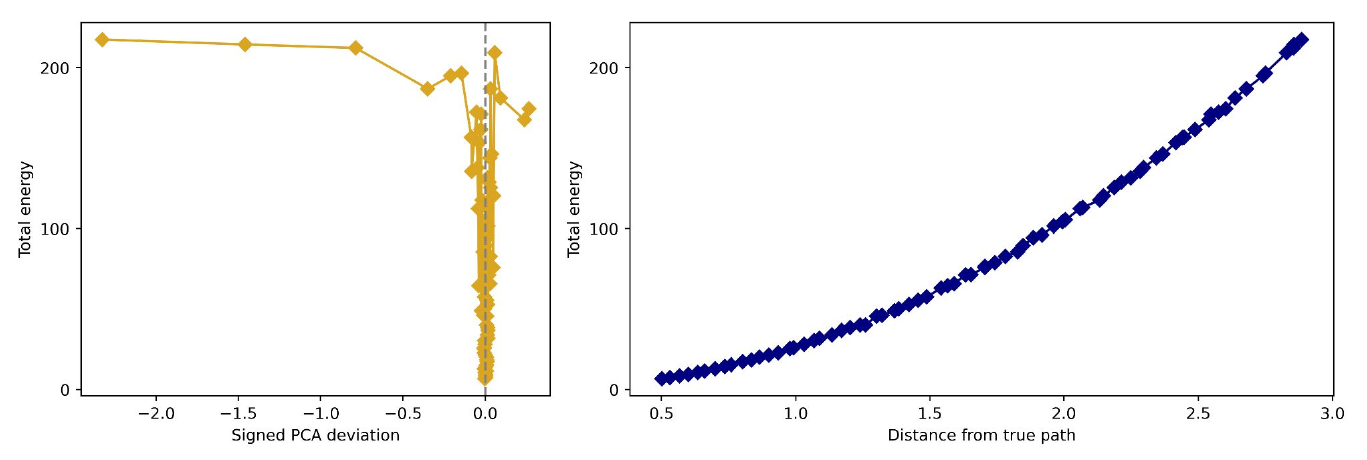}
  \caption{We plot (left) the signed PCA deviation and (right) the energy of the functional against the $L^2$ discrete distance from the true trajectory as in Theorem 1. We use a high dimension here $(d=150)$.}
\label{fig:energy_thm1_highdim}
\end{figure}}

\subsection{Optimization with closed form flow maps}

Loss functions in the token space across the trajectories can generally be cast as a calculus of variations problem under the important caveat that an optimal trajectory is under search. This is because a (continuous) loss minimization task is effectively minimizing a functional subject to the flow map and manifold constraints. We are specifically trying to minimize a loss such that the trajectory itself garners some sense of optimality. Since a loss achieves a critical point at satisfaction, which complements an Euler-Lagrange equation, our methods are compatible with general machine learning techniques in optimization on the token flow maps.

\vspace{2mm}

We will also consider more generalized calculus of variations problems, primarily motivated by \cite{dathathri2020plugplaylanguagemodels},
\begin{align}
\min \int L(x,\dot{x}) dt \ \ \ \ \ \text{subject to} \ \ \ \ \ \dot{x}  = \mathcal{P}_{x_i(t)}^{\perp} ( \frac{1}{Z_{\beta,\mu}(x_i(t))} \int e^{\beta \langle x_i(t),y \rangle} y d\mu(t,y) ), x(0) = p, \|x(t)\|=1 .
\end{align}
Here, we have automatically restricted $x$ to be a flow map. Thus, we are optimizing the functional over all possible flow maps that satisfy the constraints, as opposed to any flow map such as in Theorem.

\vspace{2mm}

We invent some new methodology of when our methods are applicable.

\vspace{2mm}

Suppose we are in search of a specific token and its trajectory that is meant to satisfy a certain cost. Suppose we are in search of an exact token $x(0) = x^*(0)$ that is meant to satisfy a certain property. For example, suppose we want to "select" an exact token by a loss function of the form
\begin{align}
\min_{x} \Big\| x(0) - x^*(0) \Big\|_2^2 .
\end{align}
Since each token has a corresponding, predetermined path, this loss function is equivalent to the discrete formulation
\begin{align}
\sum_i \Big\| \frac{x_{i+1} - x_i}{\Delta t} - \mathcal{P}_{x_i}^{\perp}( \nabla \Phi(x_i) ) \Big\|_2^2 + \frac{1}{2} \Big\| x_0 - x_0^* \Big\|_2^2 .
\end{align}
Equivalently, this loss has a functional formulation
\begin{align}
\int_0^T \Big\| \dot{x} - \mathcal{P}_{x^*(t)}(\nabla \Phi(x^*)) \Big\|_2^2 + \frac{1}{2T} \Big\| x(0) - x^*(0)\Big\|_2^2 dt  .
\end{align}
Thus, solving this loss function can be cast as an Euler-Lagrange equation, and solving the loss function for such an optimal token is equivalent to solving the Euler-Lagrange equation.

\vspace{2mm}

More generally, we can impose a more generalized constraint on the token to be selected. Let $F^*: [0,T] \rightarrow \mathbb{R}^d$ be an optimality condition and let $F : \mathbb{S}^{d-1} \subseteq \mathbb{R}^d \rightarrow \mathbb{R}^d$. We are interested in
\begin{align}
\min_{x} \Big\| F(x(0)) - F^*(0) \Big\|_2^2 .
\end{align}
Then the calculus of variations problem is
\begin{align}
& \int_0^T \Big\| \frac{d}{dt} F(x(t)) - \frac{d}{dt}F^*(t) \Big\|_2^2 + \frac{1}{2T} \Big\| F(x(0)) - F^*(0)\Big\|_2^2 dt  
\\
& = \int_0^T \Big\|  \dot{F}^T(x(t)) ( \underbrace{ \mathcal{P}_{x(t)}(\nabla \Phi(x(t))) }_{ = \dot{x} } ) - \dot{F}^*(t) \Big\|_2^2 + \frac{1}{2T} \Big\| F(x(0)) - F^*(0)\Big\|_2^2 dt .
\end{align}
Is is beneficial to include the left term instead of just taking the norm $\|F(x(t)) - F^*(t)\|$ so the Euler-Lagrange equation is well-defined. 

\vspace{2mm}

We make some notes. First, $F$ denotes any function that composes the above as a loss. It does not need to be a neural network, but it can be if it is conciliatory for the training to have the desired effect. There is another challenge. $x$ is not a neural network, it is a token that corresponds to input data. Thus, we provide a solution so that the above is learnable and well-defined. 

\vspace{2mm}

To make the above learnable, we take an elementary neural network, which is effectively a parameter that depends on time (so its only input is time) $\theta : [0,T] \rightarrow \mathbb{R}^d, \theta \in H^1([0,T]; \mathbb{R}^d \setminus \{0\})$ such that
\begin{align}
\frac{\theta_x}{\|\theta_x\|_2} = x .
\end{align}
We use this notation to mean each radial $\theta$ corresponds to a token (not necessarily seen in training, but really some point along the hypersphere). Now, we optimize over the normalized $\theta$, which is now learnable. Thus, the integrated Lagrangian is
\begin{align}
\int_0^T \Big\|  \dot{F}^T(\frac{\theta(t)}{\|\theta(t)\|_2}) ( \mathcal{P}_{\frac{\theta}{\|\theta\|_2}}(\nabla \Phi(\frac{\theta(t)}{\|\theta(t)\|_2}))  ) - \dot{F}^*(t) \Big\|_2^2 + \frac{1}{2T} \Big\| F(\frac{\theta(0)}{\|\theta(0)\|_2}) - F^*(0)\Big\|_2^2 dt  .
\end{align}
Thus, the critical $\theta/\|\theta\|_2$ is the correct token trajectory.

\begin{tcolorbox}[colback=white, colframe=BlueGreen,
  title=\textbf{Theorem (adjoint/optimal control formulation for the Transformer)}]
\noindent\textbf{Theorem 2.}
\textit{ Consider the optimization problem evaluating a continuous-depth Transformer
on the unit hypersphere $\mathbb{S}^{d-1}$
\begin{align}
\begin{cases}
& \inf_{p \in \mathbb{S}^{d-1}} \mathcal{A}[x]
  = \int_{0}^{T} L(t, x(t), \dot{x}(t)) dt
\\
& \text{subject to } \dot{x}(t) = \mathcal{X}[\mu(t)](x(t)), \ \ x(0) = p,
\end{cases}
\end{align}
where $\mathcal{X}$ is the projected attention vector field. By compactness, a solution exists. Furthermore, a necessary condition
for optimality at $p^*$ is the boundary condition
\begin{align}
  \mathcal{P}_{p^*}^{\perp}\left(\lambda(0)
  + \frac{\partial L}{\partial \dot{x}}(0)\right) &= 0,
\end{align}
where $\mathcal{P}_{x}^{\perp}$ denotes the orthogonal projection onto the
tangent space $T_{x}\mathbb{S}^{d-1}$, and the adjoint state
$\lambda(t) \in T_{x(t)}\mathbb{S}^{d-1}$ satisfies the backward
differential equation
\begin{align}
  \mathcal{P}_{x(t)}^{\perp} \left( \dot{\lambda}(t) \right)
  &= \mathcal{P}_{x(t)}^{\perp}\left(
      \frac{\partial L}{\partial x}
      - \frac{d}{dt}\left(\frac{\partial L}{\partial \dot{x}}\right)
      - \left(D_x \mathcal{X}\big|_{x(t)}\right)^{*} \lambda(t)
    \right)
\end{align}
with terminal transversality condition
$\mathcal{P}_{x(T)}^{\perp}\left(\lambda(T)
+ \frac{\partial L}{\partial \dot{x}}(T)\right) = 0$.
Here, $D_x \mathcal{X}\big|_{x(t)}$ is the Jacobian of the vector field,
${}^{*}$ is the adjoint operator, and the partial derivatives of $L$
are evaluated at $(t, x(t), \dot{x}(t))$.}
\end{tcolorbox}

\vspace{2mm}

\textit{Proof.} We introduce a Lagrange multiplier on the functional $\lambda(t)$ to enforce the constraint $\dot{x}(t) = \mathcal{X}[\mu(t)](x(t))$, yielding the functional
\begin{align}
  \mathcal{J}[x, p]
  &= \int_{0}^{T} \Bigl(
      L(t, x(t), \dot{x}(t))
      + \langle \lambda(t), \dot{x}(t) - \mathcal{X}[\mu(t)](x(t)) \rangle
    \Bigr) dt.
\end{align}
We take the first variation of $\mathcal{J}$ with respect to a perturbation $\delta x(t)$. Because $x(t)$ is constrained to $\mathbb{S}^{d-1}$, the
variation $\delta x(t)$ must lie in the tangent space $T_{x(t)}\mathbb{S}^{d-1}$
\begin{align}
  \delta \mathcal{J} 
  &= \left. \frac{d}{d\epsilon} \mathcal{J}[x + \epsilon \delta x, p] \right|_{\epsilon=0} \\
  &= \int_{0}^{T} \frac{d}{d\epsilon} \Bigl(
      L(t, x + \epsilon \delta x, \dot{x} + \epsilon \delta \dot{x}) 
      + \langle \lambda, \dot{x} + \epsilon \delta \dot{x} - \mathcal{X}[\mu(t)](x + \epsilon \delta x) \rangle
    \Bigr) \Bigg|_{\epsilon=0} dt
    \\
  &= \int_{0}^{T} \left(
      \left\langle \frac{\partial L}{\partial x}, \delta x \right\rangle
      + \left\langle \frac{\partial L}{\partial \dot{x}}, \delta \dot{x} \right\rangle
      + \langle \lambda, \delta \dot{x} \rangle
      - \left\langle \lambda, \left.\frac{d}{d\epsilon} \mathcal{X}[\mu(t)](x + \epsilon \delta x)\right|_{\epsilon=0} \right\rangle
    \right) dt
    \\
  &= \int_{0}^{T} \left(
      \left\langle \frac{\partial L}{\partial x}, \delta x \right\rangle
      + \left\langle \frac{\partial L}{\partial \dot{x}}, \delta \dot{x} \right\rangle
      + \langle \lambda, \delta \dot{x} \rangle
      - \langle \lambda, D_x \mathcal{X} \delta x \rangle
    \right) dt.
\end{align}
Rearranging terms,
\begin{align}
  \delta \mathcal{J}
  &= \int_{0}^{T} \left(
      \left\langle \frac{\partial L}{\partial x}
        - (D_x \mathcal{X})^{*} \lambda, \delta x \right\rangle
      + \left\langle \frac{\partial L}{\partial \dot{x}} + \lambda,
        \delta \dot{x} \right\rangle
    \right) dt.
\end{align}
Applying integration by parts,
\begin{align}
& \int_{0}^{T} \left\langle
    \frac{\partial L}{\partial \dot{x}} + \lambda, \dot{\delta x}
  \right\rangle dt
  = \left\langle
      \frac{\partial L}{\partial \dot{x}} + \lambda, \delta x
    \right\rangle \Bigg|_{0}^{T}
    - \int_{0}^{T} \left\langle
        \frac{d}{dt}\left(\frac{\partial L}{\partial \dot{x}}\right)
        + \dot{\lambda}, \delta x
      \right\rangle dt  \\
  &= \left\langle
      \frac{\partial L}{\partial \dot{x}}(T) + \lambda(T), \delta x(T)
    \right\rangle
    - \left\langle
        \frac{\partial L}{\partial \dot{x}}(0) + \lambda(0), \delta p
      \right\rangle 
    - \int_{0}^{T} \left\langle
        \frac{d}{dt}\left(\frac{\partial L}{\partial \dot{x}}\right)
        + \dot{\lambda}, \delta x
      \right\rangle dt,
\end{align}
where we have used the initial condition $\delta x(0) = \delta p$.
Substituting this back into the first variation and factoring out $\delta x$
\begin{align}
  \delta \mathcal{J}
  &= \left\langle
      \lambda(T) + \frac{\partial L}{\partial \dot{x}}(T), \delta x(T)
    \right\rangle
    - \left\langle
        \lambda(0) + \frac{\partial L}{\partial \dot{x}}(0), \delta p
      \right\rangle  \\
  &\quad + \int_{0}^{T} \left\langle
      \frac{\partial L}{\partial x}
      - \frac{d}{dt}\left(\frac{\partial L}{\partial \dot{x}}\right)
      - (D_x \mathcal{X})^{*} \lambda
      - \dot{\lambda},\; \delta x
    \right\rangle dt.
\end{align}
For $x(t)$ to be an optimal trajectory, the variation $\delta \mathcal{J}$
must vanish for all admissible variations $\delta x(t)$ and $\delta p$.
Setting the integrand to zero for all tangent variations $\delta x$ requires the residual gradient to vanish upon projection onto the tangent space, yielding the adjoint ODE
\begin{align}
  \mathcal{P}_{x(t)}^{\perp} \left( \dot{\lambda}(t) \right)
  &= \mathcal{P}_{x(t)}^{\perp}\left(
      \frac{\partial L}{\partial x}
      - \frac{d}{dt}\left(\frac{\partial L}{\partial \dot{x}}\right)
      - (D_x \mathcal{X})^{*} \lambda(t)
    \right).
\end{align}
Because the terminal state $x(T)$ is free, $\delta x(T)$ is arbitrary within the tangent space, so setting the $T$-boundary term to zero yields the transversality condition
\begin{align}
  \mathcal{P}_{x(T)}^{\perp}\left(
    \lambda(T) + \frac{\partial L}{\partial \dot{x}}(T)
  \right) &= 0.
\end{align}
Finally, the variation with respect to the initial token embedding $p$ leaves
the term evaluated at $t = 0$. Because $p \in \mathbb{S}^{d-1}$, the
variation $\delta p$ is constrained to $T_{p}\mathbb{S}^{d-1}$. For this
inner product to vanish for all valid $\delta p$, the vector $\lambda(0) + \frac{\partial L}{\partial \dot{x}}(0)$ must be orthogonal to the tangent space at $p^*$, meaning its projection onto the tangent space must be zero
\begin{align}
  \mathcal{P}_{p^*}^{\perp}\left(
    \lambda(0) + \frac{\partial L}{\partial \dot{x}}(0)
  \right) &= 0.
\end{align}
This completes the proof.

$ \square$

\vspace{2mm}

Theorem 2 is not in literature exactly verbatim to our knowledge, although it is based on classical calculus of variations on manifolds and control theory, and it is mostly a known result well-founded in literature \cite{lee2015globalformulationslagrangianhamiltonian} \cite{Evans2024Control} \cite{deng2024existenceoptimalpairsoptimal} \cite{Agrachev2004} \cite{Kipka2014}.

\section{The projected Euler-Lagrange equation}
\label{sec:euler-lagrange}

In this and the following sections, we develop the projected Euler-Lagrange equation \cite{Evans2024Control} and subsidiary results. The Euler-Lagrange equation on manifolds is a well-studied problem \cite{calc_of_var_on_manifolds} \cite{Vallejo_2021}.

\vspace{2mm}

\textbf{Proposition 1 (as in \cite{calc_of_var_on_manifolds}).} \textit{Consider the functional
\begin{align}
\mathcal{A}[x] = \int_{[0,T]} L(t, x(t), \dot{x}(t) ) dt
\end{align}
be a time-dependent Lagrangian such that $L : \mathbb{R}^+ \times  \mathbb{S}^{d-1} \times T \mathbb{S}^{d-1} \rightarrow \mathbb{R}$. Here, $x \in C^{\infty}([0,T])$ is a smooth curve in the tangent space beginning at $p$. Then $x$ is a critical point of the functional if
\begin{align}
\frac{\partial L}{\partial x^k} (t, x, \dot{x} ) - \frac{d}{dt} ( \frac{\partial L}{\partial \dot{x}^k}(t, x, \dot{x} ) ) = 0 .
\end{align}}

\vspace{2mm}

Note that this Euler-Lagrange equation does not have initial and terminal constraints. Any path along the manifold that satisfies this is classified as an Euler-Lagrange equation. The Euler-Lagrange equation itself as a differential equation does not encode boundary conditions.

\begin{tcolorbox}[colback=white, colframe=BlueGreen, title=\textbf{Theorem (projected Euler-Lagrange equation)}]
\textbf{Theorem 3.} \textit{Consider the classical calculus of variations problem on the unit hypersphere
\begin{align}
\begin{cases}
\inf_{x \in \mathcal{C}^1([0,T], \mathbb{R}^d)} \ \mathcal{A}[x] = \int_{[0,T]} L(t, x(t), \dot{x}(t) ) dt
\\
\text{subject to} \ \ x(0) = p, \ x(T) = q, \ \|x(t)\|=1
\end{cases}
\end{align}
equipped with a Lagrangian $L : \mathbb{R}^+ \times \mathbb{R}^d \times \mathbb{R}^d \rightarrow \mathbb{R}^+$, where $L \in C^{\infty}$. Assuming a minimizing trajectory $x^*(t)$ exists, then because the path variations $\delta x(t)$ in the tangent space $T_{x^*(t)}\mathbb{S}^{d-1}$ are arbitrary, the optimal trajectory satisfies the standard projected Euler-Lagrange equation exactly
\begin{align}
\mathcal{P}_{x^*(t)}^{\perp} \Big( \frac{\partial L}{\partial x}(t,x^*(t),\dot{x}^*(t)) -\frac{d}{dt} \frac{\partial L}{\partial \dot{x}}(t,x^*(t),\dot{x}^*(t)) \Big) = 0 .
\end{align}}
\end{tcolorbox}

\vspace{2mm}

Note that an exemplary reference that discusses this Theorem is \cite{lee2017global}, and our Theorem is presented as a Proposition in chapter 4. This reference does not provide a proof. This Theorem is also discussed in \cite{lee2015globalformulationslagrangianhamiltonian}, and a proof is provided in the case that $d=3$; however, our proof is significantly more involved and also uses more foundational theory. From our understanding, these two references are the quality sources for the exact Theorem we provide, but our work is meaningful since it is the most foundational, generalized, and specific for the Transformer. We refer to \cite{massa2012constrainedvariationalcalculussecond} \cite{Krupka_2018} \cite{2017} for various other references on Euler-Lagrange equations on manifolds.

\section{$\epsilon$-error results with geodesic balls for additively separable Lagrangians}
\label{sec:error_geodesic_ball_section}

\textbf{Theorem 4.} \textit{ Consider the calculus of variations problem
\begin{align}
\begin{cases}
\mathcal{A}[x] = \int_{[0,T]} L(t, x(t), \dot{x}(t) ) dt
\\
p \in P = \Big\{ p \in \mathbb{S}^{d-1} : \mathcal{A}[x_p] \leq \inf_q  \mathcal{A}[x_q] + \epsilon \Big\} 
\\
\text{where } x_p(t) \text{ is subject to } x_p(0) = p , \ \ \|x_p(t)\|=1 ,
\end{cases}  
\end{align}
equipped with a Lagrangian $L : \mathbb{R}^+ \times \mathbb{R}^d \times \mathbb{R}^d \rightarrow \mathbb{R}^+$, where $L \in C^{\infty}$. Let $x^*$ be the critical path. Let $L$ be such that $L(t, x, v) = F(t, x) + G(t, v)$, where $F$ is $\mu$-strongly convex in $x$, and assume suitable smoothness, Fréchet differentiability, and measurability conditions. Make the necessary assumptions on $v,\eta,p,z$ for all vectors $v$, where $\eta = \exp_{x^*}(\nu h) - x^*$ is a perturbation such that $h+x^*, h \in T_{x^*}(\mathbb{S}^{d-1})$ is parallel transported to $x^*$, $p(t) = \frac{\partial L}{\partial \dot{y}}$, $y:= x^* + \gamma h$, and $z$ is the chord between $x, x^*$. Let $\epsilon \in \mathbb{R}^+$. Let $\Sigma_r = \{ p \in \mathbb{S}^{d-1} : d_g(p, p^*) \leq r \}$ be a geodesic ball of radius $r$ centered at the optimum. Suppose we require the average of the ball to satisfy the bound
\begin{align}
\frac{1}{\text{Area}(\Sigma_r)} \int_{\Sigma_r} \Big( \mathcal{A}[x_p] - \mathcal{A}[x^*] \Big) d\mathcal{H}^{d-1}(p) \leq \epsilon .
\end{align}
Then $\text{Area}(\Sigma_r) > 0$, and the maximum radius $r$ of the geodesic ball satisfies
\begin{align}
r \leq \min \Bigg\{ R, \sqrt{\frac{2 \epsilon}{ \mu (\sin R / R)^{d-2} (\frac{d-1}{d+1}) - 2 M } }  \Bigg\},
\end{align}
assuming the above is well-defined (which it usually is for reasonable conditions on $L$). Here $M$ is a uniform upper bound dependent on the third derivatives of $L$ and $\dot{x}$. Moreover, $R$ is a fixed number that can be set beforehand.}

\vspace{2mm}

\textit{Remark.} It can be noted we have not specified a fixed terminal condition. This is not actually needed to satisfy an Euler-Lagrange equation, which we will use.

\vspace{2mm}

\textit{Remark.} One could argue the first part about positive measure is trivial due to smoothness. In this regard, our argument is rigorous and constructed in such a way that we do not have solutions occurring at irregularly often intervals, say across densely-scattered points. For example, the Lebesgue measure across the rationals is zero.

\vspace{2mm}

\textit{Remark.} Suitable assumptions include
\begin{align}
-\frac{1}{T}  \langle \eta, \frac{\partial^2 L}{\partial \dot{x}^2} \dot{\eta} \rangle \Bigg|_{t=0}^{t=T}  +   \mu \| \dot{\eta} \|_2^2  \geq 0 , \ \ \ \ \ \langle p(T), z(T) \rangle \geq \langle p(0), z(0) \rangle .
\end{align}

\vspace{2mm}

\textit{Proof.} Our proof strategy will be to integrate over a parallel-transported perturbation $h$ in the tangent space and to study the following:
\begin{align}
\label{eqn:fundamental_theorem_banach}
\mathcal{F}[x] = \mathcal{F}[x^*] + \int_0^{1} \delta \mathcal{F}[\underbrace{x^* + \gamma h}_{\displaystyle :=y}; h] d\gamma \leq \mathcal{F}[x^*] + \epsilon..
\end{align}
This follows from the fundamental theorem of calculus for Banach spaces. We provide evidence of this more closely. Define the family of curves $X_{\gamma}(t) = x^* + \gamma h$. Notice $X_0 = x^*$, $X_1 = x^* + h = x$. Here, $x^*, x$ are restricted to the manifold, and so
\begin{align}
h = \text{chord between } x \text{ and } x^*.
\end{align}
Since our Lagrangian is defined over all Euclidean space, we will study the ambient variation. Thus, it is acceptable that this family of curves is off the manifold. As long as the fundamental theorem of calculus for Banach spaces holds, considering a family of curves defined using a scaled chord (hence off the manifold) is acceptable.
 
\vspace{2mm}
Our notation is equivalent to
\begin{align}
h = x - x^* = \exp_{x^*}\bigl(\log_{x^*}(x)\bigr) - x^*, \qquad x, x^* \in \mathbb{S}^{d-1}.
\end{align}
Here $\exp_x$ is the exponential map. Now, using the Gateaux derivative,
\begin{align}
\frac{d}{d\gamma} \mathcal{F}[x^* + \gamma h]
= \lim_{\epsilon \to 0^+} \frac{\mathcal{F}[x^* + \gamma h + \epsilon h] - \mathcal{F}[x^* + \gamma h]}{\epsilon}
= \delta \mathcal{F}[x^* + \gamma h].
\end{align}
Integrating gives the result upon choosing $h = x - x^*$. Now, observe the left-hand side of \eqref{eqn:fundamental_theorem_banach} always holds, but we need to search for the collection of $x$ that satisfy the right-hand side inequality. We can integrate across
\begin{align}
\Sigma = \Bigl\{ x(0) : \mathcal{F}[x] \leq \inf_q \mathcal{F}[x_q] + \epsilon, x(t)\big|_{t=0} = x(0) \Bigr\}
\end{align}
again with the existing integral term and normalize across the domain, arriving at
\begin{align}
\frac{1}{\text{Area}(\Sigma)} \int_{\Sigma} \int_0^{1} \delta \mathcal{F}[x^* + \gamma h; h] d\gamma d\mathcal{H}^{d-1}(x(0)) \leq \epsilon.
\end{align}
Note that we need to normalize by area. This is because we are interested in
\begin{align}
\int_{\Sigma} \int_0^{1} \delta \mathcal{F}[x^* + \gamma h; h] d\gamma d\mathcal{H}^{d-1}(x(0))
\leq \int_{\Sigma} \epsilon d\mathcal{H}^{d-1}(x(0))
= \epsilon\text{Area}(\Sigma).
\end{align}
However, this formulation also necessitates $\text{Area}(\Sigma) > 0$, which we do not immediately have. Instead, we will first show there exists a $\Sigma$ satisfying the above and
\begin{align}
\int_{\Sigma} \int_0^{1} \Bigl|\delta \mathcal{F}[x^* + \gamma h; h]\Bigr| d\gamma d\mathcal{H}^{d-1}(x(0)) > 0.
\end{align}
This statement implies $\text{Area}(\Sigma) > 0$, which gives the nonzero measure condition (recall that an integral over a set of measure zero is also zero). We have defined the Hausdorff measure as
\begin{align}
\mathcal{H}^{d-1}(\Sigma)
= \liminf_{\delta \to 0} \Bigl\{ \sum_i \bigl(\text{diam}(\Sigma_i)\bigr)^k
: \Sigma \subset \bigcup_i \Sigma_i, \text{diam}(\Sigma_i) < \delta \Bigr\}.
\end{align}
We proceed with the proof of nonzero measure. Our first step is to examine the first variation. Recall we have designed our problem so that $h$ has a closed form, as does our input function $y$. We compute the first variation as in \ref{sec:first_variation_calculation}, and drop the boundary terms since they are negative. After taking the absolute value and integrating, we obtain
\begin{align}
0 \stackrel{!}{<}
\int_{[0,T]} \int_0^1 \int_{\Sigma}
\Bigl|\Bigl\langle x - x^*,
\frac{\partial L}{\partial y} - \frac{d}{dt}\frac{\partial L}{\partial \dot{y}}
\Bigr\rangle\Bigr|
d\mathcal{H}^{d-1}(y(0)) d\gamma dt
\leq \epsilon.
\end{align}
We need to show existence of a $\Sigma$ satisfying the inequality on the left. Our strategy is to pick a large ball fitting inside the respective non-Euclidean geometry and show it satisfies what we need. We attempt a construction of $\Sigma$ as the set
\begin{align}
\Sigma = \Bigl\{ x(0) \in \mathbb{S}^{d-1} : d_g\bigl(x(0), x^*(0)\bigr) < r \Bigr\}.
\end{align}
We have denoted by $d_g$ the geodesic distance. We will determine $r$ later. All we need to show is that there exists an $x(0)$, sufficiently far from the optimal $x^*(0)$ yet contained in this set, that is not optimal. The result of the left inequality follows immediately by smoothness.
 
\vspace{2mm}
We make some additional remarks. Note that in the above, $y$ is allowed to be off the manifold, and $x - x^*$ need not lie in the tangent space. We will need both of these facts for the result of Theorem~3 to apply. These strategies are valid as long as $L$ is defined in the extrinsic space as well, which we have permitted. In particular, in the first variation computation as in Theorem~3, we did not yet invoke $\|x\| = 1$ nor $h \in T_{x}\mathbb{S}^{d-1}$ to reach that point.
 
\vspace{2mm}
Since we are integrating across the geodesic ball at $t = 0$, we consider all $x$ that start in such a geodesic ball. Let
\begin{align}
x = \exp_{x^*}(\nu h) \in \mathbb{S}^{d-1},
\end{align}
where $\nu$ is a small constant such that $x(0) \in \Sigma$, and $h(t) \in C^{\infty}$ is a suitable test function in the tangent space. \textit{Note that we are redefining} $h$ \textit{from what we used earlier.} We now have
\begin{align}
y = x^* + \gamma\bigl(\exp_{x^*}(\nu h) - x^*\bigr)
\end{align}
as our new variational function. We restrict $h$ so that $\bigl\|\tfrac{d}{dt}\bigl(\exp_{x^*}(\nu h) - x^*\bigr)\bigr\|_2 \leq Cr$. In general, we can take $h$ to be independent of time in the parallel-transported sense (we do not literally take $h$ independent of time, since this would imply $h$ does not remain in the tangent space, making the exponential map ill-defined). We leave $h$ time-dependent for generalization purposes. We must show
\begin{align}
\Bigl|\Bigl\langle \bigl(\exp_{x^*}(\nu h) - x^*\bigr),
\frac{\partial L}{\partial\bigl(x^* + \gamma(\exp_{x^*}(\nu h) - x^*)\bigr)}
- \frac{d}{dt}\frac{\partial L}{\partial\bigl(x^* + \gamma(\exp_{x^*}(\nu h) - x^*)\bigr)'}
\Bigr\rangle\Bigr| \neq 0.
\end{align}
Suppose for the sake of contradiction that $x$ (and hence $y$) is a critical point of the calculus of variations problem. If $x$ does not satisfy this, the result is immediate. Examining $t = 0$ simplifies the problem greatly and does not affect the result by smoothness. After cancellation of terms, we are left with
\begin{align}
\Bigl|\Bigl\langle \bigl(\exp_{x^*}(\nu h(0)) - x^*(0)\bigr),
\Bigl(\frac{\partial L}{\partial(x^* + \gamma(\exp_{x^*}(\nu h) - x^*))}
- \frac{d}{dt}\frac{\partial L}{\partial(x^* + \gamma(\exp_{x^*}(\nu h) - x^*))'}\Bigr)\Bigr|_{t=0}
\Bigr\rangle\Bigr|
\stackrel{?}{\neq} 0.
\end{align}
Note that the Euler-Lagrange equation is not satisfied exactly due to the perturbation; thus both terms in the inner product are nonzero. We are therefore taking a directional derivative with a vector in the same direction, so they are not orthogonal, and
\begin{align}
\Bigl|\Bigl\langle \bigl(\exp_{x^*}(\nu h(0)) - x^*(0)\bigr),
\Bigl(\frac{\partial L}{\partial(x^* + \gamma(\exp_{x^*}(\nu h) - x^*))}
- \frac{d}{dt}\frac{\partial L}{\partial(x^* + \gamma(\exp_{x^*}(\nu h) - x^*))'}\Bigr)\Bigr|_{t=0}
\Bigr\rangle\Bigr|
\neq 0.
\end{align}
By smoothness, we have constructed our desired $\Sigma$, completing the first part of the proof.
 
\vspace{2mm}

We proceed with the proof of the bound. We find a bound on the region around the critical path:
\begin{align}
&\frac{1}{T \times \text{Area}(\Sigma)}
\int_{[0,T]} \int_0^1 \int_{\Sigma}
\langle (\exp_{x^*}(\nu h) - x^*),
(\frac{\partial L}{\partial(x^* + \gamma(\exp_{x^*}(\nu h) - x^*))}  \\
&\hspace{5cm}
- \frac{d}{dt}\frac{\partial L}{\partial(x^* + \gamma(\exp_{x^*}(\nu h) - x^*))'})
|_{(t,y,\dot{y})}
\rangle
d\mathcal{H}^{d-1}(y(0)) d\gamma dt  \\[1em]
&= \frac{1}{T \times \text{Area}(\Sigma)}
\int_{[0,T]} \int_0^1 \int_{\Sigma}
\frac{1}{\gamma}
\langle \gamma(\exp_{x^*}(\nu h) - x^*),
(\frac{\partial L}{\partial(x^* + \gamma(\exp_{x^*}(\nu h) - x^*))}  \\
&\hspace{5cm}
- \frac{d}{dt}\frac{\partial L}{\partial(x^* + \gamma(\exp_{x^*}(\nu h) - x^*))'})
|_{(t,y,\dot{y})}
\rangle
d\mathcal{H}^{d-1}(y(0)) d\gamma dt.
\end{align}
We have used the notation
\begin{align}
\frac{\partial L}{\partial(x^* + \gamma(\exp_{x^*}(\nu h) - x^*))'}
|_{(t,y,\dot{y})}
= \frac{\partial L}{\partial \dot{x}}(t, y, (x^* + \gamma(\exp_{x^*}(\nu h) - x^*))'),
\end{align}
noting these are the same quantity; we adopt this notation for brevity and to emphasize that the derivative is taken with respect to the correct function. Our integral with respect to $y(0)$ is well-defined because our Euler-Lagrange equation is evaluated along this path. From now on we omit the argument for simplicity. We use our unconstrained functional because the path is free, so there are no rigid ODE multiplier terms.

\vspace{2mm}

We proceed by inserting zero and using the $\mu$-strong convexity condition for separately additive Lagrangians. It is important to note that this inner product is \emph{not} taken with absolute value. We get
\begin{align}
&\frac{1}{T \times \text{Area}(\Sigma)}
\int_{[0,T]} \int_0^1 \int_{\Sigma}
\frac{1}{\gamma}
\langle \gamma(\exp_{x^*}(\nu h) - x^*) - x^*(t) + x^*(t),
\\
&\hspace{1cm}
\frac{\partial L}{\partial(x^* + \gamma(\exp_{x^*}(\nu h) - x^*))}
- \frac{d}{dt}\frac{\partial L}{\partial(x^* + \gamma(\exp_{x^*}(\nu h) - x^*))'}
\rangle
d\mathcal{H}^{d-1}(y(0)) d\gamma dt
\\[1em]
&= \frac{1}{T \times \text{Area}(\Sigma)}
\int_{[0,T]} \int_0^1 \int_{\Sigma}
\frac{1}{\gamma}
\langle x^*(t) + \gamma(\exp_{x^*}(\nu h) - x^*) - x^*(t),
\\
&\hspace{1cm}
\frac{\partial L}{\partial(x^* + \gamma(\exp_{x^*}(\nu h) - x^*))}
- \frac{d}{dt}\frac{\partial L}{\partial(x^* + \gamma(\exp_{x^*}(\nu h) - x^*))'}
+ \frac{\partial L}{\partial x^*} - \frac{\partial L}{\partial x^*}
\rangle
d\mathcal{H}^{d-1}(y(0)) d\gamma dt
\\[1em]
&= \frac{1}{T \times \text{Area}(\Sigma)}
\int_{[0,T]} \int_0^1 \int_{\Sigma}
\frac{1}{\gamma}
[
\langle x^*(t) + \gamma(\exp_{x^*}(\nu h) - x^*) - x^*(t),
\frac{\partial L}{\partial(x^* + \gamma(\exp_{x^*}(\nu h) - x^*))} - \frac{\partial L}{\partial x^*}\rangle  \\
&\hspace{1cm}
+ \langle x^*(t) + \gamma(\exp_{x^*}(\nu h) - x^*) - x^*(t),
-\frac{d}{dt}\frac{\partial L}{\partial(x^* + \gamma(\exp_{x^*}(\nu h) - x^*))'}
+ \frac{\partial L}{\partial x^*}\rangle
]
d\mathcal{H}^{d-1}(y(0)) d\gamma dt 
\\[1em]
&\geq \frac{1}{T \times \text{Area}(\Sigma)}
\int_{[0,T]} \int_0^1 \int_{\Sigma}
\frac{1}{\gamma}
[
\mu\|x^*(t) + \gamma(\exp_{x^*}(\nu h) - x^*) - x^*(t)\|_2^2 
\\
&\hspace{1cm}
+ \langle x^*(t) + \gamma(\exp_{x^*}(\nu h) - x^*) - x^*(t),
-\frac{d}{dt}\frac{\partial L}{\partial(x^* + \gamma(\exp_{x^*}(\nu h) - x^*))'}
+ \frac{\partial L}{\partial x^*}\rangle
]
d\mathcal{H}^{d-1}(y(0)) d\gamma dt  \\[1em]
&= \frac{1}{T \times \text{Area}(\Sigma)}
\int_{[0,T]} \int_0^1 \int_{\Sigma}
\frac{1}{\gamma}
[ \mu\|\gamma(\exp_{x^*}(\nu h) - x^*)\|_2^2 
\\
&\hspace{1cm}
+ \langle \gamma(\exp_{x^*}(\nu h) - x^*),
-\frac{d}{dt}\frac{\partial L}{\partial(x^* + \gamma(\exp_{x^*}(\nu h) - x^*))'}
+ \frac{\partial L}{\partial x^*}\rangle
]
d\mathcal{H}^{d-1}(y(0)) d\gamma dt.
\end{align}
The inequality follows from additively-separable $\mu$-strong convexity. Setting the shorthand
\begin{align}
\eta = \exp_{x^*}(\nu h(t)) - x^*
\end{align}
and performing a Taylor expansion gives
\begin{align}
\frac{\partial L}{\partial(\dot{x}^* + \gamma\dot{\eta})}(t, x^* + \gamma\eta, \dot{x}^* + \gamma\dot{\eta})
&= \frac{\partial L}{\partial \dot{x}}(t, x^*, \dot{x}^*)
+ \gamma(\frac{\partial^2 L}{\partial x\partial \dot{x}}\eta
+ \frac{\partial^2 L}{\partial \dot{x}^2}\dot{\eta})  \\
&\quad + \frac{\gamma^2}{2}(
\eta^T\frac{\partial^3 L}{\partial x^2\partial \dot{x}}\eta
+ 2\eta^T\frac{\partial^3 L}{\partial x\partial \dot{x}^2}\dot{\eta}
+ \dot{\eta}^T\frac{\partial^3 L}{\partial \dot{x}^3}\dot{\eta}
) + \mathcal{O}(\gamma^3).
\end{align}
Hence,
\begin{align}
&\frac{1}{T \times \text{Area}(\Sigma)}
\int_{[0,T]} \int_0^1 \int_{\Sigma}
\frac{1}{\gamma}
[\mu\|\gamma(\exp_{x^*}(\nu h) - x^*)(t)\|_2^2  \\
&\hspace{1.5cm}
+ \langle \gamma(\exp_{x^*}(\nu h) - x^*)(t),
-\frac{d}{dt}\frac{\partial L}{\partial(x^* + \gamma(\exp_{x^*}(\nu h) - x^*))'}
+ \frac{\partial L}{\partial x^*}\rangle
]
d\mathcal{H}^{d-1}(y(0)) d\gamma dt  \\[1em]
&= \frac{1}{T \times \text{Area}(\Sigma)}
\int_{[0,T]} \int_0^1 \int_{\Sigma}
\frac{1}{\gamma}
[\mu\|\gamma(\exp_{x^*}(\nu h) - x^*)(t)\|_2^2  \\
&\hspace{1.5cm}
+ \langle \gamma(\exp_{x^*}(\nu h) - x^*)(t),
-\frac{d}{dt}(\frac{\partial L}{\partial \dot{x}}(t,x^*,\dot{x}^*)
+ \gamma(\frac{\partial^2 L}{\partial x\partial \dot{x}}\eta
+ \frac{\partial^2 L}{\partial \dot{x}^2}\dot{\eta}) + \mathcal{O}(\gamma^2))
+ \frac{\partial L}{\partial x^*}\rangle
]
d\mathcal{H}^{d-1}(y(0)) d\gamma dt  \\[1em]
&= \frac{1}{T \times \text{Area}(\Sigma)}
\int_{[0,T]} \int_0^1 \int_{\Sigma}
\frac{1}{\gamma}
[\mu\|\gamma(\exp_{x^*}(\nu h) - x^*)(t)\|_2^2  \\
&\hspace{1.5cm}
+ \langle \gamma(\exp_{x^*}(\nu h) - x^*)(t),
-\frac{d}{dt}\frac{\partial L}{\partial \dot{x}}(t,x^*,\dot{x}^*) + \frac{\partial L}{\partial x^*}\rangle  \\
&\hspace{1.5cm}
+ \langle \gamma(\exp_{x^*}(\nu h) - x^*)(t),
-\gamma\frac{d}{dt}(\frac{\partial^2 L}{\partial x\partial \dot{x}}\eta
+ \frac{\partial^2 L}{\partial \dot{x}^2}\dot{\eta})
+ \frac{d}{dt}\mathcal{O}(\gamma^2)\rangle
]
d\mathcal{H}^{d-1}(y(0)) d\gamma dt.
\end{align}
The remainder of our proof strategy is as follows. The first term is clearly positive and poses no problem. Our approach with the second term requires care due to the manifold constraint. The third term will be simplified by assumptions.

\vspace{2mm}

Because $x^*$ is a critical path constrained to $\mathbb{S}^{d-1}$, it satisfies the \emph{projected} Euler-Lagrange equation. Therefore, we cannot immediately simplify this term using the fact $x^*$ is critical. This means the unconstrained Euler-Lagrange equation is parallel to the normal vector $x^*(t)$
\begin{align}
-\frac{d}{dt}\frac{\partial L}{\partial \dot{x}}(t,x^*,\dot{x}^*) + \frac{\partial L}{\partial x^*} = \lambda(t)x^*(t)
\end{align}
for some scalar Lagrange multiplier $\lambda(t)$. Taking the inner product with our ambient perturbation $\gamma \eta = \gamma(\exp_{x^*}(\nu h) - x^*)$, we get
\begin{align}
\int_{[0,T]} \int_0^1 \int_{\Sigma}
\langle \gamma(\exp_{x^*}(\nu h) - x^*), \lambda(t)x^*(t) \rangle
d\mathcal{H}^{d-1}(y(0)) d\gamma dt.
\end{align}
Since $\exp_{x^*}(\nu h) = \cos(\nu\|h\|)x^* + \sin(\nu\|h\|)\frac{h}{\|h\|}$ and $\langle h, x^* \rangle = 0$, the total inner product with $x^*$ included simplifies to $\lambda(t)(\cos(\nu\|h\|) - 1)$. The chord $\eta$ has a normal component, so this does not vanish. However, since the perturbation lies within a geodesic ball of radius $r$, we have $1 - \cos(\nu\|h\|) \leq \frac{1}{2}r^2$. Thus, this geometric residual is bounded below by $-\frac{1}{2}\|\lambda\|_{\infty}r^2$.

\vspace{2mm}

To simplify the third term, by assumption on $L$, we use
\begin{align}
\eta^{\top} \frac{d}{dt}\frac{\partial^2 L}{\partial x \partial \dot{x}} \eta = 0
\end{align}
vanishes by the separately additive condition. Observe this does not contradict $h(0)$ lying in the geodesic ball; $h$ is still free within the geodesic ball at $t = 0$. We apply our conditions and integrating by parts. Let us bundle the $\mathcal{O}(\gamma^2)$ Taylor terms and the $\mathcal{O}(r^k)$ Euler-Lagrange normal residuals into an $R(\gamma, t)$. Our integral reduces to
\begin{align}
&\frac{1}{T \cdot \text{Area}(\Sigma)} \int_{[0,T]} \int_0^1 \int_{\Sigma} \frac{1}{\gamma} \Bigg[ \mu\|\gamma(\exp_{x^*}(\nu h) - x^*)\|_2^2 \\
&\hspace{1.5cm} - \langle \gamma \eta, \gamma\frac{d}{dt}\Big(\frac{\partial^2 L}{\partial \dot{x}^2}\dot{\eta}\Big) \rangle + \gamma \lambda(t)(\cos(\nu\|h\|) - 1) + \langle \gamma \eta, R(\gamma, t) \rangle \Bigg] d\mathcal{H}^{d-1}(y(0)) d\gamma dt \\
&= \frac{1}{T \cdot \text{Area}(\Sigma)} \int_{[0,T]} \int_0^1 \int_{\Sigma} \Bigg[ \gamma \mu\|\exp_{x^*}(\nu h) - x^*\|_2^2
\\
&\hspace{1.5cm} + \gamma \Big( - \frac{1}{T}\langle \eta, \frac{\partial^2 L}{\partial \dot{x}^2}\dot{\eta}\rangle\Big|_{t=0}^{t=T} + \langle \dot{\eta}, \frac{\partial^2 L}{\partial \dot{x}^2}\dot{\eta}\rangle \Big)
\\
&\hspace{1.5cm} + \lambda(t)(\cos(\nu\|h\|) - 1) +  \mathcal{O}(\gamma^2 r^3) \Bigg] d\mathcal{H}^{d-1}(y(0)) d\gamma dt 
\\
&\geq \frac{1}{T \cdot \text{Area}(\Sigma)} \int_{[0,T]} \int_0^1 \int_{\Sigma} \Bigg[ \gamma \mu \|\exp_{x^*}(\nu h) - x^*\|_2^2 - \frac{1}{2}\|\lambda\|_{\infty}r^2 - C\gamma^2 r^3 \Bigg] d\mathcal{H}^{d-1}(y(0)) d\gamma dt .
\end{align}
Let us bound $\|\lambda \|_{\infty}$, integrate, and combine the residual terms with a uniform bound $M$. It can be noted $\gamma$ is small which helps keep $M$ small, so it is also of interest to keep $\| \lambda \|_{\infty}$ small. We obtain
\begin{align}
&\frac{1}{T \cdot \text{Area}(\Sigma)}
\int_{[0,T]} \int_0^1 \int_{\Sigma}
[\gamma\mu \|\exp_{x^*}(\nu h) - x^*\|_2^2
- r^2 ( \frac{1}{2} \widetilde{C}_1 + C \gamma^2 r )]
d\mathcal{H}^{d-1}(y(0)) d\gamma dt 
\\
&\geq \frac{1}{T \cdot \text{Area}(\Sigma)}
\int_{[0,T]} \int_0^1
[\gamma\mu V_{\Sigma}
- M r^2 \text{Area}(\Sigma)]
dt d\gamma \\[0.5em]
&\geq \frac{1}{2\text{Area}(\Sigma)}\mu V_{\Sigma}\big|_{t=0} - Mr^2.
\end{align}
Our argument uses the fact that
\begin{align}
\frac{1}{T}\int_0^T f dt \geq \inf f.
\end{align}
Our lower bound is independent of time, which will ultimately allow us to examine $t = 0$. And we are almost done. Also, we know
\begin{align}
\frac{1}{2\text{Area}(\Sigma)}\mu V_{\Sigma}\big|_{t=0} - Mr^2 \leq \epsilon .
\end{align}
The evaluation at $t = 0$ is justified because we can always choose $h$ to be constant in time (i.e., parallel transported), so at least one solution exists satisfying what we need. Using a geodesic-ball polar-coordinate integral,
\begin{align}
\label{eqn:polar}
\frac{V_{\Sigma}\big|_{t=0}}{\text{Area}(\Sigma)}
= \frac{\int_0^{r} s^2 A(s) ds}{\int_0^{r} A(s) ds}
\geq \frac{|\mathbb{S}^{d-2}|\int_0^r s^2\sin^{d-2}(s) ds}{|\mathbb{S}^{d-2}|\int_0^r \sin^{d-2}(s) ds}
\geq \left(\frac{\sin r}{r}\right)^{d-2}\frac{d-1}{d+1}r^2,
\end{align}
using $\sin(s) \geq [(\sin r)/r]s$ and $\sin(s) \leq s$. On the unit sphere, the squared chord length is exactly $2 - 2\cos(s)$. The Taylor expansion is
\begin{align}
2 - 2\cos(s) = s^2 - \frac{s^4}{12} + \mathcal{O}(s^6) ,
\end{align}
therefore sufficiently truncate this to get the $s^2 A$ term in \ref{eqn:polar}. Thus we need
\begin{align}
\frac{1}{2}\mu \left(\frac{\sin r}{r}\right)^{d-2}\frac{d-1}{d+1}r^2 - Mr^2  \leq \epsilon.
\end{align}
It is highly nontrivial to isolate $r$ above. Instead, we impose a maximum on $r$, denoted $R$. Substituting this in, and using $\sin R/R \leq \sin r/r$,
\begin{align}
\frac{1}{2}\mu \left(\frac{\sin R}{R}\right)^{d-2}\frac{d-1}{d+1}r^2 - Mr^2  \leq \epsilon.
\end{align}
Choosing
\begin{align}
\Sigma = \left\{
x \in \mathbb{S}^{d-1} : d_g(x, x^*(0))
\leq \min\left\{
R,
\sqrt{\frac{2\epsilon}{\mu \left(\tfrac{\sin R}{R}\right)^{d-2}\tfrac{d-1}{d+1} - 2M}}
\right\}
\right\},
\end{align}
and we are done.

$\square$

\section{$\epsilon$-ball result for fixed path and varying initial point}

\textbf{Theorem 5.} \textit{Consider the optimization problem
\begin{align}
\begin{cases}
\displaystyle\inf_{p \in \mathbb{S}^{d-1}} \mathcal{A}[x] = \int_{0}^{T} \mathcal{L}(t, x, \dot{x}, \lambda)  dt
\\
\mathcal{L}(t, x, \dot{x}, \lambda) = L(t, x, \dot{x}) + \langle \lambda(t), \dot{x}(t) - \mathcal{X}(x(t)) \rangle
\\[6pt]
\text{subject to} \ \dot{x}(t) = \mathcal{X}[\mu(t)]\left(x(t)\right), \ \ x(0) = p.
\end{cases}
\end{align}
Let $p^*$ be the optimal initial token embedding. Assume $\mathcal{X}$ is uniformly Lipschitz continuous in $x$. We define the sub-optimal boundary
condition set as
\begin{align}
P = \left\{ p \in \mathbb{S}^{d-1} : \left\| \mathcal{P}_{p}^{\perp}\left(\lambda_p(0)
+ \frac{\partial L}{\partial \dot{x}_p}(0)\right) \right\|_2 \leq \epsilon \right\} ,
\end{align}
where $x_p(t)$ is the forward-propagated trajectory initiated at $x_p(0) = p$, and
$\lambda_p(t)$ is the corresponding adjoint state. Assume the Lagrangian $L \in C^\infty$ is jointly $\mu$-strongly convex in its state and velocity arguments. Let
$\eta(t) = x_p(t) - x^*(t)$ be the path variation initiated by $\eta(0) = p - p^*$. Let
$\epsilon \in \mathbb{R}^+$. Then the Hausdorff (Lebesgue) measure of $P$ is positive.
Moreover, the set $P$ bounds a geodesic ball of radius $r$, where
\begin{align}
r \leq \min \Bigg\{ R, \frac{\epsilon}{[ \mu c_T - MR] C_R (\sin R / R)^{d-2}\left(\dfrac{d-1}{d+1}\right) } \Bigg\} ,
\end{align}
assuming the denominator is positive. Here $M$ is a uniform upper bound dependent on the higher derivatives of $L$ and the ODE flow Lipschitz constant, $R$ is a fixed radius set beforehand, $ c_T = \frac{1 - e^{-2 L_f T}}{2 L_f} $, and $C_R = \left(\frac{\sin(R/2)}{R/2}\right)^2$.}

\vspace{2mm}

\textit{Proof.} Our strategy is to integrate the inner product of the boundary condition over a geodesic ball
and use the first variation of the action functional. Define the operator
\begin{align}
\mathcal{G}(p) = \mathcal{P}_{p}^{\perp}\left(\lambda_p(0)
+ \frac{\partial L}{\partial \dot{x}_p}(0)\right).
\end{align}
We prove via the remark that
\begin{align}
\label{eqn:riemannian_grad}
\langle \mathcal{G}(p), \eta(0) \rangle
= \int_0^T \left[
\Bigl\langle \eta(t), \frac{\partial L}{\partial x}(y, \dot{y}) \Bigr\rangle
+ \Bigl\langle \dot{\eta}(t), \frac{\partial L}{\partial \dot{x}}(y, \dot{y}) \Bigr\rangle
\right] dt.
\end{align}
We construct our integration region. Let $\Sigma \subset P$ be a geodesic ball of radius $r$ centered at $p^*$, such that
\begin{align}
\Sigma = \Bigl\{ p \in \mathbb{S}^{d-1} : d_g(p, p^*) < r \Bigr\}.
\end{align}
We integrate the inner product across $\Sigma$ and normalize by the area. On the left side,
we apply the Cauchy--Schwarz inequality. Because we are inside $P$,
$\|\mathcal{G}(p)\|_2 \leq \epsilon$. On the ball, $\|\eta(0)\|_2 = \|p - p^*\|_2 \leq r$.
Thus
\begin{align}
\frac{1}{\mathrm{Area}(\Sigma)} \int_{\Sigma} \langle \mathcal{G}(p), \eta(0) \rangle 
d\mathcal{H}^{d-1}(p)
\leq \frac{1}{\mathrm{Area}(\Sigma)} \int_{\Sigma} \epsilon \|\eta(0)\|_2 
d\mathcal{H}^{d-1}(p) \leq \epsilon r.
\end{align}
Let us insert zero and apply the $\mu$-strong convexity condition
\begin{align}
&\frac{1}{\mathrm{Area}(\Sigma)} \int_{\Sigma} \int_0^T \left[
\Bigl\langle \eta, \frac{\partial L}{\partial x}(y, \dot{y})
- \frac{\partial L}{\partial x}(x^*, \dot{x}^*)
+ \frac{\partial L}{\partial x}(x^*, \dot{x}^*) \Bigr\rangle \right.  \\
&\hspace{3.5cm} \left.
+ \Bigl\langle \dot{\eta}, \frac{\partial L}{\partial \dot{x}}(y, \dot{y})
- \frac{\partial L}{\partial \dot{x}}(x^*, \dot{x}^*)
+ \frac{\partial L}{\partial \dot{x}}(x^*, \dot{x}^*) \Bigr\rangle
\right] dt  d\mathcal{H}^{d-1}(p)  \\[1em]
&= \frac{1}{\mathrm{Area}(\Sigma)} \int_{\Sigma} \int_0^T \left[
\Bigl\langle \eta, \frac{\partial L}{\partial x}(y, \dot{y})
- \frac{\partial L}{\partial x}(x^*, \dot{x}^*) \Bigr\rangle
+ \Bigl\langle \dot{\eta}, \frac{\partial L}{\partial \dot{x}}(y, \dot{y})
- \frac{\partial L}{\partial \dot{x}}(x^*, \dot{x}^*) \Bigr\rangle
\right] dt  d\mathcal{H}^{d-1}(p)  \\
&\quad + \frac{1}{\mathrm{Area}(\Sigma)} \int_{\Sigma}
\underbrace{\int_0^T \left[
\Bigl\langle \eta, \frac{\partial L}{\partial x}(x^*, \dot{x}^*) \Bigr\rangle
+ \Bigl\langle \dot{\eta}, \frac{\partial L}{\partial \dot{x}}(x^*, \dot{x}^*) \Bigr\rangle
\right] dt}_{=\langle \mathcal{G}(p^*),\eta(0)\rangle}
d\mathcal{H}^{d-1}(p).
\end{align}
Because $p^*$ is optimal, the necessary boundary condition is satisfied, so
$\mathcal{G}(p^*) = 0$. This annihilates the entire second integral.

\vspace{2mm}

We perform a Taylor expansion. We can note
\begin{align}
 \begin{pmatrix} \frac{\partial L}{\partial x}(y, \dot{y}) - \frac{\partial L}{\partial x}(x^*, \dot{x}^*) \\ \frac{\partial L}{\partial \dot{x}}(y, \dot{y}) - \frac{\partial L}{\partial \dot{x}}(x^*, \dot{x}^*) \end{pmatrix} = \begin{pmatrix} \frac{\partial^2 L}{\partial x^2} & \frac{\partial^2 L}{\partial x \partial \dot{x}} \\ \frac{\partial^2 L}{\partial \dot{x} \partial x} & \frac{\partial^2 L}{\partial \dot{x}^2} \end{pmatrix} \begin{pmatrix} \eta \\ \dot{\eta} \end{pmatrix} + \mathcal{O}(\|\eta\|^2) .
\end{align}
We get
\begin{align}
& = \frac{1}{\mathrm{Area}(\Sigma)} \int_{\Sigma} \int_0^T \left[
\begin{pmatrix} \eta \\ \dot{\eta} \end{pmatrix}^{\top}
\begin{pmatrix}
\dfrac{\partial^2 L}{\partial x^2} & \dfrac{\partial^2 L}{\partial x \partial \dot{x}} \\[6pt]
\dfrac{\partial^2 L}{\partial \dot{x} \partial x} & \dfrac{\partial^2 L}{\partial \dot{x}^2}
\end{pmatrix}
\begin{pmatrix} \eta \\ \dot{\eta} \end{pmatrix}
+ \mathcal{O}(\|\eta\|^3)
\right] dt  d\mathcal{H}^{d-1}(p)  \\
&\geq \frac{1}{\mathrm{Area}(\Sigma)} \int_{\Sigma} \int_0^T \left[
\mu\|\eta\|_2^2 + \mu\|\dot{\eta}\|_2^2 + \mathcal{O}(\|\eta\|^3)
\right] dt  d\mathcal{H}^{d-1}(p)  \\
&\geq \frac{1}{\mathrm{Area}(\Sigma)} \int_{\Sigma} \int_0^T \left[
\mu\|\eta\|_2^2 + \mathcal{O}(\|\eta\|^3)
\right] dt  d\mathcal{H}^{d-1}(p).
\end{align}
The second line follows by the $\mu$-strong convexity and the third follows from dropping a positive term. By the Lipschitz continuity of the ODE flow, the trajectories are bounded below such that via Gronwall's inequality
\begin{align}
\int_0^T \|\eta(t)\|_2^2 dt \geq \|\eta(0)\|_2^2 \int_0^T e^{-2 L_f t} dt = \|\eta(0)\|_2^2 \underbrace{ \left( \frac{1 - e^{-2 L_f T}}{2 L_f}  \right) }_{= c_T} .
\end{align}
Let us bound the $\mathcal{O}(\| \eta \|^3)$ term, and allow $\|\eta(0)\| \leq r$. Since we assumed $\mathcal{X}$ is Lipschitz continuous, by Grönwall's inequality, we get $\|\eta(t)\|_2^2 \geq \|\eta(0)\|_2^2 e^{-2L_f t}$ and $ \|\eta(t)\|_2 \leq e^{L_f t} \|\eta(0)\|_2 \leq e^{L_f t} r$. We bound by a constant $M$
\begin{align}
\geq \frac{1}{\mathrm{Area}(\Sigma)} \int_{\Sigma} \left[
\mu c_T \|\eta(0)\|_2^2 - Mr\|\eta(0)\|_2^2
\right] d\mathcal{H}^{d-1}(p).
\end{align}
Using the polar integral definition
$V_{\Sigma} = \int_{\Sigma}\|\eta(0)\|_2^2  d\mathcal{H}^{d-1}(p)$, we get
\begin{align}
= \left( \mu  c_T - Mr \right) \frac{V_{\Sigma}}{\mathrm{Area}(\Sigma)}.
\end{align}
Connecting to our left-hand Cauchy-Schwarz bound
\begin{align}
\epsilon r \geq \left( \mu c_T - Mr \right) \frac{V_{\Sigma}}{\mathrm{Area}(\Sigma)}.
\end{align}
We substitute the known geodesic-ball polar-coordinate lower bound for
$\frac{V_{\Sigma}}{\mathrm{Area}(\Sigma)}$. Because $\|\eta(0)\|_2$ is the chord, we consider
$C_R = \left(\frac{\sin(R/2)}{R/2}\right)^2$
\begin{align}
\epsilon r \geq \left( \mu c_T - Mr \right) C_R
\Bigl(\frac{\sin r}{r}\Bigr)^{d-2} \frac{d-1}{d+1}  r^2.
\end{align}
We divide by $r$. It is nontrivial to isolate $r$, so we use the fact that $\sin R/R \leq \sin r/r$ for $r \leq R$
\begin{align}
\epsilon \geq r \Bigl(\frac{\sin R}{R}\Bigr)^{d-2} C_R
\frac{d-1}{d+1} \left[ \mu c_T   - MR \right].
\end{align}
Isolating $r$, we arrive at our final bound 
\begin{align}
r \leq \frac{\epsilon}{[ \mu c_T - MR] C_R (\sin R / R)^{d-2}\left(\dfrac{d-1}{d+1}\right) } ,
\end{align}
and taking the $\text{min}$ with $R$ gives the result.

$\square$

\vspace{2mm}

\textit{Remark.} Equation \ref{eqn:riemannian_grad} follows since by definition of the Gateaux derivative,
\begin{align}
\delta \mathcal{A} = \int_0^T \left[ \Bigl\langle \eta(t), \frac{\partial L}{\partial x} \Bigr\rangle + \Bigl\langle \dot{\eta}(t), \frac{\partial L}{\partial \dot{x}} \Bigr\rangle \right] dt .
\end{align}
Introducing the ODE constraint of the adjoint state,
\begin{align}
\delta \mathcal{A} = \int_0^T \left[ \Bigl\langle \eta, \frac{\partial L}{\partial x} \Bigr\rangle + \Bigl\langle \dot{\eta}, \frac{\partial L}{\partial \dot{x}} \Bigr\rangle + \Bigl\langle \lambda, \dot{\eta} - \frac{\partial \mathcal{X}}{\partial x}\eta \Bigr\rangle \right] dt  + \mathcal{O}(\|\eta\|^2),
\end{align}
which is inserting zero since the dynamics constraint is satisfied. We drop the error for simplicity. Thus, by regrouping, we see
\begin{align}
\delta \mathcal{A} = \int_0^T \left[ \Bigl\langle \frac{\partial L}{\partial x} - \left(\frac{\partial \mathcal{X}}{\partial x}\right)^\top \lambda,  \eta \Bigr\rangle + \Bigl\langle \frac{\partial L}{\partial \dot{x}} + \lambda,  \dot{\eta} \Bigr\rangle \right] dt .
\end{align}
Via integration by parts,
\begin{align}
\int_0^T \Bigl\langle \frac{\partial L}{\partial \dot{x}} + \lambda,  \dot{\eta} \Bigr\rangle dt = \left[ \Bigl\langle \frac{\partial L}{\partial \dot{x}} + \lambda,  \eta \Bigr\rangle \right]_0^T - \int_0^T \Bigl\langle \frac{d}{dt}\left(\frac{\partial L}{\partial \dot{x}} + \lambda\right),  \eta \Bigr\rangle dt .
\end{align}
Thus,
\begin{align}
\delta \mathcal{A} = \left[ \Bigl\langle \lambda + \frac{\partial L}{\partial \dot{x}},  \eta \Bigr\rangle \right]_0^T + \int_0^T \left\langle \frac{\partial L}{\partial x} - \left(\frac{\partial \mathcal{X}}{\partial x}\right)^\top \lambda - \frac{d}{dt}\left(\lambda + \frac{\partial L}{\partial \dot{x}}\right),  \eta \right\rangle dt .
\end{align}
Define the dynamics of the adjoint state by 
\begin{align}
\dot{\lambda} = \frac{\partial L}{\partial x} - \left(\frac{\partial \mathcal{X}}{\partial x}\right)^\top \lambda - \frac{d}{dt}\frac{\partial L}{\partial \dot{x}}.
\end{align}
The integral is zero and so what remains is the tranversality condition
\begin{align}
\lambda(T) + \frac{\partial L}{\partial \dot{x}}(T) = 0 .
\end{align}
We are left with
\begin{align}
\delta \mathcal{A} = - \Bigl\langle \lambda(0) + \frac{\partial L}{\partial \dot{x}}(0),  \eta(0) \Bigr\rangle .
\end{align}
By definition, the Euclidean gradient of the action with respect to the initial state is the vector $g$ that satisfies $\delta \mathcal{A} = \langle g, \eta(0) \rangle$. Thus, in the ambient Euclidean space $\nabla_p \mathcal{A} = \lambda(0) + \frac{\partial L}{\partial \dot{x}}(0)$. Equivalently,
\begin{align}
\langle \mathcal{G}(p), \eta(0) \rangle = \int_0^T \left[ \Bigl\langle \eta(t), \frac{\partial L}{\partial x}(y, \dot{y}) \Bigr\rangle + \Bigl\langle \dot{\eta}(t), \frac{\partial L}{\partial \dot{x}}(y, \dot{y}) \Bigr\rangle \right] dt .
\end{align}

\section{Discretization results under closed form flow maps}

\textbf{Theorem 6.} \textit{ Consider the functional
\begin{align}
\mathcal{A}[h] = \int_0^T (  \frac{1}{2} \|\dot{h}(t)\|_2^2 - \langle \dot{h}(t), \mathcal{P}_{x(t)}^{\perp} ( \nabla_x \log ( \int e^{\beta \langle x(t), y \rangle} d\mu(t,y) )  ) \rangle + \frac{1}{2T} \| h(0) - x(0) \|_2^2 ) dt .
\end{align}
Consider the functional discretized
\begin{align}
\widetilde{\mathcal{A}}_{\Delta t}[h] = \Delta t \sum_i (  \frac{1}{2} \|\frac{h_{i+1} - h_i}{\Delta t} \|_2^2 - \langle \frac{h_{i+1} - h_i}{\Delta t} , \mathcal{P}_{x_{i+ \frac{1}{2}}}^{\perp} ( \nabla_x \log ( \int e^{\beta \langle x_{i+\frac{1}{2}}, y \rangle} d\mu_{i+\frac{1}{2}}(y) )  ) \rangle + \frac{1}{2T} \| h_0 - x_0 \|_2^2 )  .
\end{align}
Suppose
\begin{align}
\| \dot{h} \|_2 \leq M_1, \| \ddot{h} \|_2 \leq M_2, \|\dddot{h} \|_2 \leq M_3 .
\end{align}
Then
\begin{align}
\Bigg| \mathcal{A}[h] - \widetilde{\mathcal{A}}_{\Delta t}[h] \Bigg| = \mathcal{O}((\Delta t)^2)
\end{align}
for all $h$.}

\vspace{2mm}

\textit{Proof.} We know from the midpoint quadrature rule \cite{rimmer_midpoint}
\begin{align}
& \int_{t_i}^{t_i + \Delta t} (  \frac{1}{2} \|\dot{h}(t)\|_2^2 - \langle \dot{h}(t), \mathcal{P}_{x(t)}^{\perp} ( \nabla_x \log ( \int e^{\beta \langle x(t), y \rangle} d\mu(t,y) )  ) \rangle + \frac{1}{2T} \| h(0) - x(0) \|_2^2 ) dt 
\\[2em]
& = \Delta t  (  \frac{1}{2} \|\dot{h}(t_i + \frac{\Delta t}{2})\|_2^2 - \langle \dot{h}(t_i + \frac{\Delta t}{2}), \mathcal{P}_{x(t_i + \frac{\Delta t}{2})}^{\perp} ( \nabla_x \log ( \int e^{\beta \langle x(t_i + \frac{\Delta t}{2}), y \rangle} d\mu(t_i + \frac{\Delta t}{2},y) )  ) \rangle  + \frac{1}{2T} \| h(0) - x(0) \|_2^2 )
\\
& \ \ \ \ \ \   - \frac{(\Delta t)^3}{24} \frac{d^2}{dt^2}(  \frac{1}{2} \|\dot{h}(t)\|_2^2 - \langle \dot{h}(t), \mathcal{P}_{x(t)}^{\perp} ( \nabla_x \log ( \int e^{\beta \langle x(t), y \rangle} d\mu(t,y) )  ) \rangle ) \Bigg|_{t=\xi} .
\end{align}
The proof follows by summing over the intervals that form $[0,T]$, and taking the upper bound on the second derivative. Thus, it remains to find this upper bound. Notice
\begin{align}
& \frac{d^2}{dt^2}(  \frac{1}{2} \|\dot{h}(t)\|_2^2 - \langle \dot{h}(t), \underbrace{ \mathcal{P}_{x(t)}^{\perp} ( \nabla_x \log ( \int e^{\beta \langle x(t), y \rangle} d\mu(t,y) )  ) }_{=P(t)} \rangle ) 
\\
& = \frac{d}{dt} ( \langle \dot{h}, \ddot{h} \rangle - \langle \ddot{h}, P \rangle - \langle \dot{h}, \dot{P} \rangle ) 
\\
& =  \frac{d}{dt} ( \langle \dot{h} - P, \ddot{h} \rangle  - \langle \dot{h}, \dot{P} \rangle )
\\
& = \langle \ddot{h} - \dot{P}, \ddot{h} \rangle + \langle \dot{h} - P,  \dddot{h} \rangle - \langle \ddot{h}, \dot{P} \rangle - \langle \dot{h}, \ddot{P} \rangle 
\\
& = \| \ddot{h} \|_2^2  - \langle \dot{P}, \ddot{h} \rangle + \langle \dot{h} - P,  \dddot{h} \rangle - \langle \ddot{h}, \dot{P} \rangle - \langle \dot{h}, \ddot{P} \rangle 
\\
& \leq \| \ddot{h} \|_2^2  +  \| \dot{P} \|_2 \| \ddot{h} \|_2 + ( \|\dot{h}\|_2 + \| P\|_2)  \| \dddot{h} \|_2 + \| \ddot{h} \|_2 \| \dot{P} \|_2 + \| \dot{h} \|_2 \| \ddot{P} \|_2
\\
& \leq M_2^2  +  \| \dot{P} \|_2 M_2 + ( M_1 + \| P\|_2)  M_3+ M_2 \| \dot{P} \|_2 + M_1 \| \ddot{P} \|_2 .
\end{align}
Now, we find bounds on the derivatives of $P$. Observe
\begin{align}
\mathcal{P}_{x(t)}^{\perp} ( \nabla_x \log ( \int e^{\beta \langle x(t), y \rangle} d\mu(t,y) ) ) = (I - x(t) x(t)^T)  ( \nabla_x \log ( \int e^{\beta \langle x(t), y \rangle} d\mu(t,y) ) )  .
\end{align}
Taking the first derivative,
\begin{align}
& \frac{d}{dt} (I - x(t) x(t)^T)  ( \nabla_x \log ( \int e^{\beta \langle x(t), y \rangle} d\mu(t,y) )  )
\\
& = - (\dot{x} x^T + x \dot{x}^T )  ( \nabla \Phi(x) ) + (I - x(t) x(t)^T) \frac{d}{dt} \nabla \Phi(x) .
\end{align}
Taking the second derivative
\begin{align}
& \frac{d}{dt} \Bigg[ - (\dot{x} x^T + x \dot{x}^T )  ( \nabla \Phi(x) ) + (I - x(t) x(t)^T) \frac{d}{dt} \nabla \Phi(x) \Bigg] 
\\[2em]
& = - (\ddot{x} x^T + 2\dot{x} \dot{x}^T + x \ddot{x}^T )  ( \nabla \Phi(x) ) -2 (\dot{x} x^T + x \dot{x}^T )\frac{d}{dt} \nabla \Phi(x)  + (I - x(t) x(t)^T) \frac{d^2}{dt^2} \nabla \Phi(x)  .
\end{align}
Next, we differentiate the potential. Observe
\begin{align}
\frac{d}{dt} \nabla \Phi & = \frac{d}{dt} \frac{1}{Z_{\beta}(x(t))}     \int \beta y e^{\beta \langle x(t), y \rangle} d\mu(t,y) 
\\
& = \frac{1}{Z_{\beta}^2} \Bigg[ Z \int \beta^2 y  \langle \dot{x}, y \rangle e^{\beta \langle x, y \rangle} d\mu + Z \int \beta y e^{\beta \langle x, y \rangle} \partial_t \mu - \dot{Z} \int \beta y e^{\beta \langle x, y \rangle } d\mu \Bigg] .
\end{align}
Observe the numerator is bounded since $x,y$ lie on the unit sphere, and $\dot{x}$ is also bounded. We show the denominator is bounded below. Notice
\begin{align}
Z_{\beta} = \int e^{\beta \langle x, y \rangle } d\mu  \geq e^{-\beta} .
\end{align}
Thus, the derivative is bounded. Notice the second derivative is bounded similarly, since again by the quotient rule, we have $Z_{\beta}$ in the denominator, which is bounded below, and clearly the numerator is bounded above. Thus, we have all the bounded we need. The bounds on $\|\dot{P}\|_2, \|\ddot{P}\|_2$ follow from Cauchy-Schwarz and the triangle inequality. Thus, we have shown
\begin{align}
\frac{d^2}{dt^2}(  \frac{1}{2} \|\dot{h}(t)\|_2^2 - \langle \dot{h}(t), \underbrace{ \mathcal{P}_{x(t)}^{\perp} ( \nabla_x \log ( \int e^{\beta \langle x(t), y \rangle} d\mu(t,y) )  ) }_{=P(t)} \rangle ) \leq \Theta ,
\end{align}
where $\Theta$ is some ultimate constant. From this, we can conclude
\begin{align}
& \Bigg| \int_{t_i}^{t_i + \Delta t} (  \frac{1}{2} \|\dot{h}(t)\|_2^2 - \langle \dot{h}(t), \mathcal{P}_{x(t)}^{\perp} ( \nabla_x \log ( \int e^{\beta \langle x(t), y \rangle} d\mu(t,y) )  ) \rangle + \frac{1}{2T} \| h(0) - x(0) \|_2^2 ) dt 
\\
& -  \Delta t \sum_i (  \frac{1}{2} \|\dot{h}(t_i + \frac{\Delta t}{2})\|_2^2 - \langle \dot{h}(t_i + \frac{\Delta t}{2}), \mathcal{P}_{x(t_i + \frac{\Delta t}{2})}^{\perp} ( \nabla_x \log ( \int e^{\beta \langle x(t_i + \frac{\Delta t}{2}), y \rangle} d\mu(t_i + \frac{\Delta t}{2},y) )  ) \rangle  
\\
& \ \ \ \ \ \ \ \ \ \ \ \ + \frac{1}{2T} \| h(0) - x(0) \|_2^2 )  \Bigg| \leq \frac{\Theta (\Delta t)^3}{24}  .
\end{align}
Summing up across the intervals, we conclude
\begin{align}
& \Bigg| \int_{0}^{T} (  \frac{1}{2} \|\dot{h}(t)\|_2^2 - \langle \dot{h}(t), \mathcal{P}_{x(t)}^{\perp} ( \nabla_x \log ( \int e^{\beta \langle x(t), y \rangle} d\mu(t,y) )  ) \rangle  ) dt - \sum_{i=1}^N \Delta t  (  \frac{1}{2} \|\dot{h}(t_i + \frac{\Delta t}{2})\|_2^2 
\\
& - \langle \dot{h}(t_i + \frac{\Delta t}{2}), \mathcal{P}_{x(t_i + \frac{\Delta t}{2})}^{\perp} ( \nabla_x \log ( \int e^{\beta \langle x(t_i + \frac{\Delta t}{2}), y \rangle} d\mu(t_i + \frac{\Delta t}{2},y) )  ) \rangle ) \Bigg|  = \mathcal{O}((\Delta t)^2)  ,
\end{align}
and we are done.

$ \square $

\vspace{2mm}

\textit{Remark.} Generally, it is more suitable to do a left-endpoint rule when accounting for the boundary condition at $t=0$. If this point is omitted, our flow discretization along a midpoint-type method is perfectly fine, and is actually preferred due to a better convergence rate. Here, we have chosen each layer output as a midpoint.

\textbf{\begin{figure}[htbp]
  \vspace{0mm}
  \centering
  \includegraphics[scale=0.54]{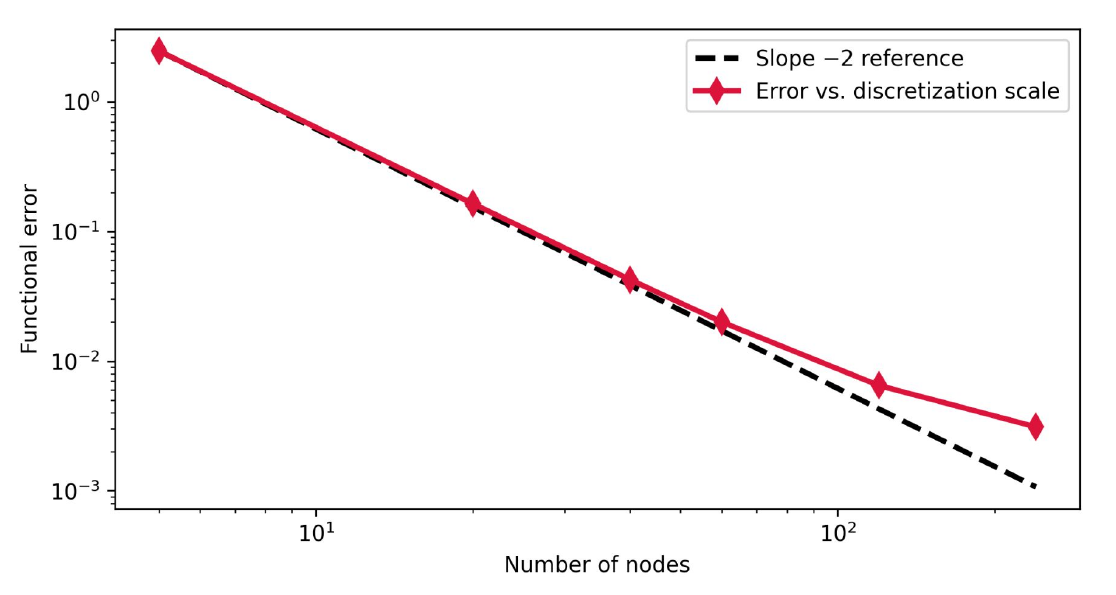}
  \caption{We provide a log-log error plot of empirical results of Theorem 6, showing that the error estimate is indeed is $\mathcal{O}((\Delta t)^2)$.}
\label{fig:energy_thm1_highdim}
\end{figure}}

\section{Energies with pushforwards}

\textbf{Theorem 7.} \textit{Consider the calculus of variations problem}
\begin{align}
\delta \int_{\mathbb{S}^{d-1}} L(x) d\mu_0(x) = 0
\quad \text{subject to} \quad
\phi_{\sharp}\mu_0 = \mu_1,\quad L \in \mathcal{C}^{\infty}(\mathbb{S}^{d-1};\mathbb{R}^+)
\end{align}
\textit{equipped with the Borel $\sigma$-algebra}
\begin{align}
\bigl(\mathbb{S}^{d-1}, \mathcal{B}, \mathcal{H}^{d-1}\bigr).
\end{align}
\textit{Here, $\phi : \mathbb{S}^{d-1} \to \mathbb{S}^{d-1}$ is the pushforward flow map induced from the optimal free paths satisfying the projected Euler-Lagrange equation. A solution to this calculus of variations problem is the point measure (Dirac mass)}
\begin{align}
\mu_0(\cdot) = \delta_p(\cdot),
\end{align}
\textit{at some optimal initial point $p \in \mathbb{S}^{d-1}$. Moreover, $\phi_{\sharp} \mu_0^* = \delta_q$.}
 
\vspace{2mm}
\textit{Proof.} Our strategy is to find a critical point of a Lagrange multiplier-type first variation problem. Using Lagrange multipliers and the pushforward-of-measure formula \cite{pushforwardmeasure2024}, the augmented functional is
\begin{align}
\mathcal{A}[\mu_0;\lambda]
&= \int_{\mathbb{S}^{d-1}} L d\mu_0
  + \int_{\mathbb{S}^{d-1}} \lambda(y) d(\phi_{\sharp}\mu_0 - \mu_1)(y)  \\
&= \int_{\mathbb{S}^{d-1}} \bigl(L(x) + \lambda(\phi(x))\bigr) d\mu_0(x)
  - \int_{\mathbb{S}^{d-1}} \lambda(y) d\mu_1(y).
\end{align}
We compute the first variation. Let $\nu$ be a signed test Borel measure with $\int_{\mathbb{S}^{d-1}} d\nu = 0$, so that the perturbed measure remains a probability measure. Then
\begin{align}
\mathcal{A}[\mu_0 + \epsilon\nu;\lambda]
= \int_{\mathbb{S}^{d-1}} \bigl(L(x) + \lambda(\phi(x))\bigr) d(\mu_0 + \epsilon\nu)(x)
  - \int_{\mathbb{S}^{d-1}} \lambda(y) d\mu_1(y).
\end{align}
Differentiating with respect to $\epsilon$,
\begin{align}
\frac{d}{d\epsilon}\mathcal{A}[\mu_0 + \epsilon\nu;\lambda]
= \int_{\mathbb{S}^{d-1}} \bigl(L(x) + \lambda(\phi(x))\bigr) d\nu(x).
\end{align}
This is already $\epsilon$-independent, so no evaluation at $\epsilon = 0$ is needed. Setting the first variation to zero,
\begin{align}
0 = \int_{\mathbb{S}^{d-1}} \bigl(L(x) + \lambda(\phi(x))\bigr) d\nu(x).
\end{align}
Since $\nu$ is an arbitrary test measure with zero total mass, the integrand must be constant on the support of the optimal measure $\mu_0$. Thus,
\begin{align}
\label{eqn:pushforward_map_first_var_result}
L(x) + \lambda(\phi(x)) = \text{const}
\qquad \mu_0\text{-a.e.}
\end{align}
We now show that a Dirac mass satisfies this condition. Take $\mu_0 = \delta_p$ for some $p \in \mathbb{S}^{d-1}$. By definition of the pushforward, $\mu_1 = \phi_{\sharp}\delta_p = \delta_{\phi(p)}$. Since the support of $\delta_p$ is the single point $\{p\}$, condition \eqref{eqn:pushforward_map_first_var_result} only needs to hold at $x = p$, giving
\begin{align}
L(p) + \lambda(\phi(p)) = \text{const}.
\end{align}
This is trivially satisfied by setting the constant $C = L(p) + \lambda(\phi(p))$. Because the first variation condition holds everywhere on the support of the measure, $\delta_p$ satisfies the necessary conditions for optimality.

$\square$
 
\vspace{2mm}
 
\textbf{Theorem 8.} \textit{Let $\mu_0 = \delta_{x_0^*}$ be an optimal Dirac mass satisfying the calculus of variations problem}
\begin{align}
\delta \int_{\mathbb{S}^{d-1}} L(x) d\mu_0(x) = 0
\quad \text{subject to} \quad
\phi_{\sharp}\mu_0 = \mu_1,\quad L \in \mathcal{C}^{\infty}(\mathbb{S}^{d-1};\mathbb{R}^+)
\end{align}
\textit{centered at $x_0^*$. Let $L$ be $\ell$-Lipschitz. Consider a collection of particles at $t = 0$ on the unit sphere $\{x_i(0)\}_{i \in [n]}$ with the empirical measure}
\begin{align}
\mu_0 = \frac{1}{n}\sum_{i=1}^n \delta_{x_i(0)}.
\end{align}
\textit{Consider the calculus of variations problem threshold}
\begin{align}
\mu_0 \in \Bigl\{ \mu \in \mathcal{P}(\mathbb{S}^{d-1}, \mathcal{B})
: \int_{\mathbb{S}^{d-1}} L d\mu \leq L(x_0^*) + \epsilon \Bigr\}.
\end{align}
\textit{If the particles $\{x_i(0)\}_i$ satisfy}
\begin{align}
\frac{1}{n}\sum_{i=1}^n d_g\bigl(x_i(0), x_0^*\bigr) \leq \frac{\epsilon}{\ell},
\end{align}
\textit{then}
\begin{align}
\int_{\mathbb{S}^{d-1}} L d\mu_0 \leq L(x_0^*) + \epsilon.
\end{align}
 
\vspace{2mm}
\textit{Proof.} We bound the integral over the empirical measure directly, using the fact that it is a finite sum of Dirac masses, and then apply the $\ell$-Lipschitz condition. Fix the optimal point $x_0^*$ and let $\{x_i(0)\}_i$ be any collection of particles whose average geodesic distance satisfies the bound. Then
\begin{align}
\int_{\mathbb{S}^{d-1}} L d\mu_0
&= \frac{1}{n}\sum_{i=1}^n L(x_i(0))  \\
&= \frac{1}{n}\sum_{i=1}^n \Bigl(L(x_0^*) + L(x_i(0)) - L(x_0^*)\Bigr)  \\
&\leq \frac{1}{n}\sum_{i=1}^n \Bigl(L(x_0^*) + \bigl|L(x_i(0)) - L(x_0^*)\bigr|\Bigr)  \\
&= L(x_0^*) + \frac{1}{n}\sum_{i=1}^n \bigl|L(x_i(0)) - L(x_0^*)\bigr|  \\
&\leq L(x_0^*) + \frac{1}{n}\sum_{i=1}^n \ell d_g\bigl(x_i(0), x_0^*\bigr)  \\
&= L(x_0^*) + \ell\frac{1}{n}\sum_{i=1}^n d_g\bigl(x_i(0), x_0^*\bigr)  \\
&\leq L(x_0^*) + \ell\cdot\frac{\epsilon}{\ell}  \\
&= L(x_0^*) + \epsilon,
\end{align}
which yields the required upper bound.

$\square$

\section{Geodesics and their errors}
\label{sec:geodesics_and_errors}

Theorem 9. \textit{Let $x^*:[0,T]\to\mathbb{S}^{d-1}$ be a constant-speed minimizing geodesic connecting $p,q$, and write $\nu=\|\dot{x}^*(t)\|_2$ (note that it is constant with respect to $t$). Let $w:[0,T]\to\mathbb{S}^{d-1}$ be a $C^2$ approximate path satisfying the same endpoint conditions and
\begin{align}
0\leq \mathcal{A}[w]-\mathcal{A}[x^*]\leq\epsilon, \qquad \mathcal{A}[w]=\int_0^T\|\dot{w}(t)\|_2^2 dt.
\end{align}
Suppose $\|\dot{w}\|_{L^\infty}\leq V$. Define the orthogonal projection $P_x=I-xx^\top, x\in\mathbb{S}^{d-1}$. Then
\begin{align}
\left| \int_0^T \left\langle w-x^*,P_{x^*}\ddot{w} \right\rangle_{\mathbb{R}^d}dt \right| \leq{}& \epsilon + \left( \frac{7}{2}\nu^2+\frac{3}{2}\nu V \right) \int_0^T\|w-x^*\|_2^2 dt. \end{align}
}

\vspace{2mm}

\textit{Proof.} Set $v(t)=w(t)-x^*(t)$. Because both paths satisfy the same endpoints $v(0)=v(T)=0$. Since $x^*$ is a constant-speed spherical geodesic, its extrinsic geodesic equation is $\ddot{x}^*=-\nu^2x^*$. In particular, $P_{x^*}\ddot{x}^*=0$. Via self-adjointness of the operator
\begin{align}
I &:= \int_0^T \langle v,P_{x^*}\ddot{w}\rangle dt = \int_0^T \langle P_{x^*}v,\ddot{v}\rangle dt. 
\end{align}
Integration by parts gives
\begin{align}
I = -\int_0^T \left\langle \frac{d}{dt}(P_{x^*}v),\dot{v} \right\rangle dt.
\end{align}
Notice that $
\frac{d}{dt}P_{x^*} = -\dot{x}^*x^{*\top} -x^*\dot{x}^{*\top}$.
Consequently,
\begin{align} I=& -\int_0^T\|P_{x^*}\dot{v}\|_2^2 dt + \int_0^T \Big[ \langle x^*,v\rangle \langle\dot{x}^*,\dot{v}\rangle + \langle\dot{x}^*,v\rangle \langle x^*,\dot{v}\rangle \Big]dt
\\
& =  -\int_0^T \left[ \|P_{x^*}\dot{v}\|_2^2 + \langle\dot{x}^*,v\rangle^2 \right]dt - \int_0^T \left[ \frac{1}{2}\|v\|_2^2\langle\dot{x}^*,\dot{v}\rangle + \langle\dot{x}^*,v\rangle\langle v,\dot{v}\rangle \right]dt .
\end{align}
We use $w$ and $x^*$ are unit vectors, $\langle x^*,v\rangle=-\frac{1}{2}\|v\|_2^2$, and differentiating $\langle x^*,\dot{v}\rangle = -\langle\dot{x}^*,v\rangle -\langle v,\dot{v}\rangle$. 
Via Cauchy-Schwarz,$
\left| \frac{1}{2}\|v\|^2\langle\dot{x}^*,\dot{v}\rangle + \langle\dot{x}^*,v\rangle\langle v,\dot{v}\rangle \right| \leq \frac{3}{2}\nu\|v\|^2\|\dot{v}\|$. It is also true that $\|\dot{v}\| \leq\|\dot{w}\|+\|\dot{x}^*\| \leq V+\nu$. Therefore,
\begin{align} 
|I| \leq{}& \int_0^T\|\dot{v}\|^2dt + \nu^2\int_0^T\|v\|^2dt + \frac{3}{2}\nu(V+\nu)\int_0^T\|v\|^2dt. \end{align}
Since $\dot{w}=\dot{x}^*+\dot{v}$, we have
\begin{align}
\mathcal{A}[w]-\mathcal{A}[x^*] = \int_0^T \left( \|\dot{v}\|^2 +2\langle\dot{x}^*,\dot{v}\rangle \right)dt.
\end{align}
Integration by parts, together with $\ddot{x}^*=-\nu^2x^*$, gives
\begin{align} 2\int_0^T\langle\dot{x}^*,\dot{v}\rangle dt &= -2\int_0^T\langle\ddot{x}^*,v\rangle dt
\\ &= 2\nu^2\int_0^T\langle x^*,v\rangle dt
\\ &= -\nu^2\int_0^T\|v\|^2dt. \end{align}
We have used $\langle x^*, v \rangle = -\frac{1}{2}\|v\|^2$. By $\epsilon$-suboptimality,
\begin{align}
\int_0^T\|\dot{v}\|^2dt \leq \epsilon+\nu^2\int_0^T\|v\|^2dt.
\end{align}
Therefore,
\begin{align}
|I| & \leq \epsilon + \left( 2\nu^2 + \frac{3}{2}\nu V + \frac{3}{2}\nu^2 \right) \int_0^T\|v\|^2 dt 
\\
& = \epsilon + \left( \frac{7}{2}\nu^2+\frac{3}{2}\nu V \right) \int_0^T\|w-x^*\|^2 dt .
\end{align}

$ \square $

\vspace{2mm}

\noindent \textit{Remark.} Now let us consider constructing a perturbation. Let smooth $h$ be in the tangent bundle, so $h(t)\in T_{x^*(t)}\mathbb{S}^{d-1}$ and $h(0)=h(T)=0$, and define such that $z=x^*+h$. Set $w=\widetilde{z}=\frac{z}{\|z\|}$. Tangency implies $\|z\|^2=1+\|h\|^2$. Now, notice
\begin{align}
\dot{w} & = \frac{\dot{z}\|z\| - z\left(\frac{d}{dt}\|z\|\right)}{\|z\|^2} = \frac{\dot{z}\|z\| - z\frac{\langle z, \dot{z} \rangle}{\|z\|}}{\|z\|^2} = \frac{\dot{z}}{\|z\|} - \frac{z\langle z, \dot{z} \rangle}{\|z\|^3}
\\
\implies  \|\dot{w}\|^2 & = \left\langle \frac{\dot{z}}{\|z\|}, \frac{\dot{z}}{\|z\|} \right\rangle - 2\left\langle \frac{\dot{z}}{\|z\|}, \frac{z\langle z, \dot{z} \rangle}{\|z\|^3} \right\rangle + \left\langle \frac{z\langle z, \dot{z} \rangle}{\|z\|^3}, \frac{z\langle z, \dot{z} \rangle}{\|z\|^3} \right\rangle
\\
& = \frac{\|\dot{z}\|^2}{\|z\|^2} - \underbrace{ \frac{\langle z, \dot{z} \rangle^2}{\|z\|^4} }_{\text{non-negative}} \leq \| \dot{z} \|^2 .
\end{align}
Recall from the setup that $z = x^* + h$, where $h$ is orthogonal to $x^*$ (tangent to the sphere). Therefore, $\|z\|^2 = \|x^*\|^2 + \|h\|^2 = 1 + \|h\|^2$. Because $1 + \|h\|^2 \geq 1$, we know that dividing by $\|z\|^2$ makes the fraction smaller. Thus, normalization cannot increase the  ambient kinetic energy $\mathcal{A}[w]\leq\mathcal{A}[z]$. Also recall we assumed $\mathcal{A}[z] \leq\mathcal{A}[x^*]+\epsilon$. Now, note
\begin{align}
\|w - x^*\|^2 & = \|w\|^2 - 2\langle w, x^* \rangle + \|x^*\|^2
\\
&  = 2 - 2\langle w, x^* \rangle
\\
\implies \langle w, x^* \rangle & = \left\langle \frac{x^* + h}{\sqrt{1 + \|h\|^2}}, x^* \right\rangle = \frac{\langle x^*, x^* \rangle + \langle h, x^* \rangle}{\sqrt{1 + \|h\|^2}}
\\
&  = \frac{1}{\sqrt{1 + \|h\|^2}}
\\
\label{eqn:h_inequality}
\implies \|w - x^*\|^2 & = 2 - \frac{2}{\sqrt{1 + \|h\|^2}} = 2\left(1 - \frac{1}{\sqrt{1+\|h\|^2}}\right) \leq \| h \|^2 ,
\end{align}
where the last inequality follows by considering $2\left(1 - \frac{1}{u}\right) \leq u^2 - 1$, and multiplying both sides by $u$ gives $2(u - 1) \leq u(u^2 - 1) \iff 2 \leq u(u + 1) \iff 2 \leq u^2 + u$ since we safely divide by $(u - 1)$. Since $u \geq 1$, we can trace the inequality back to its original state, which is true. Combining everything, mostly using the pointwise quality of \ref{eqn:h_inequality}, we get
\begin{align}
\epsilon + \left( \frac{7}{2}\nu^2+\frac{3}{2}\nu V \right) \int_0^T\|w-x^*\|^2 dt \leq \epsilon+ \left( \frac{7}{2}\nu^2+\frac{3}{2}\nu V \right) \int_0^T\|h(t)\|^2dt .
\end{align}

\bibliographystyle{plainnat}
\bibliography{bibliography}

\appendix

\section{First variation calculation}
\label{sec:first_variation_calculation}

Here, we derive our first variation as necessary in \ref{sec:error_geodesic_ball_section}. Consider the first variation definition using the functional
\begin{align}
\delta \mathcal{F}[y; h] & = \lim_{\epsilon \to 0^+} \frac{\int_0^T L(t, y + \epsilon h, \dot{y} + \epsilon \dot{h}) dt - \int_0^T L(t, y, \dot{y}) dt}{\epsilon}
\\
& = \lim_{\epsilon \to 0^+} \int_0^T \frac{L(t, y + \epsilon h, \dot{y} + \epsilon \dot{h}) - L(t, y, \dot{y})}{\epsilon} dt.
\end{align}
Under a Taylor expansion,
\begin{align}
L(t, y + \epsilon h, \dot{y} + \epsilon \dot{h}) = L(t, y, \dot{y}) + \epsilon \left\langle \frac{\partial L}{\partial y}, h \right\rangle + \epsilon \left\langle \frac{\partial L}{\partial \dot{y}}, \dot{h} \right\rangle + \mathcal{O}(\epsilon^2) .
\end{align}
Substituting back into the integrand,
\begin{align}
\delta \mathcal{F}[y; h] & = \lim_{\epsilon \to 0^+} \int_0^T \frac{\epsilon \left\langle \frac{\partial L}{\partial y}, h \right\rangle + \epsilon \left\langle \frac{\partial L}{\partial \dot{y}}, \dot{h} \right\rangle + \mathcal{O}(\epsilon^2)}{\epsilon} dt 
\\
&  = \lim_{\epsilon \to 0^+} \int_0^T \left( \left\langle \frac{\partial L}{\partial y}, h \right\rangle + \left\langle \frac{\partial L}{\partial \dot{y}}, \dot{h} \right\rangle + \mathcal{O}(\epsilon) \right) dt.
\end{align}
Taking the limit,
\begin{align}
\delta \mathcal{F}[y; h] = \int_0^T \left( \left\langle \frac{\partial L}{\partial y}, h \right\rangle + \left\langle \frac{\partial L}{\partial \dot{y}}, \dot{h} \right\rangle \right) dt .
\end{align}
Applying integration by parts gives the result
\begin{align}
\delta \mathcal{F}[y; h] = \int_0^T \left\langle h(t), \frac{\partial L}{\partial y} - \frac{d}{dt} \frac{\partial L}{\partial \dot{y}} \right\rangle dt + \left\langle \frac{\partial L}{\partial \dot{y}}(T), h(T) \right\rangle - \left\langle \frac{\partial L}{\partial \dot{y}}(0), h(0) \right\rangle .
\end{align}
In our formulation, $h(t)$ is the scaled chord $h(t) = x(t) - x^*(t)$. The boundary conditions do not initially annihilate in this formulation. We will drop the boundary terms as in our Theorem hypotheses by ensuring the two terms are collectively a positive value, thus making an upper bound of $\epsilon$ suitable, since $a+b\leq c \implies a \leq c$ when $b \geq 0$.

\vspace{2mm}

\begin{tabular}{@{}l}
  Department of Mathematics, Purdue University \\
  West Lafayette, IN 47907, United States \\
  \texttt{agracyk@purdue.edu}
\end{tabular}

\end{document}